\documentclass{article}

\usepackage[final]{corl_2019} 

\usepackage{times}
\usepackage{soul}
\usepackage{url}
\usepackage[utf8]{inputenc}
\usepackage[small]{caption}
\usepackage{graphicx}
\usepackage{amsmath}
\usepackage{booktabs}
\usepackage{subcaption}
\usepackage{amsfonts}
\usepackage{amssymb}
\usepackage{amsthm}
\usepackage[noend]{algpseudocode}
\usepackage[ruled,vlined,linesnumbered]{algorithm2e}
\usepackage{fullpage}
\usepackage[table,x11names]{xcolor}
\definecolor{lightgray2}{RGB}{234, 242, 206}
\definecolor{blue}{RGB}{225, 232, 245}
\definecolor{yellow}{RGB}{255, 255, 255}

\usepackage{array}
\usepackage{arydshln}
\usepackage{wrapfig}
\let\OLDthebibliography\thebibliography
\renewcommand\thebibliography[1]{
  \OLDthebibliography{#1}
  \setlength{\parskip}{0pt}
  \setlength{\itemsep}{1pt plus 0.1ex}
}

\newtheorem{innercustomgeneric}{\customgenericname}
\providecommand{\customgenericname}{}
\newcommand{\newcustomtheorem}[2]{%
  \newenvironment{#1}[1]
  {%
   \renewcommand\customgenericname{#2}%
   \renewcommand\theinnercustomgeneric{##1}%
   \innercustomgeneric
  }
  {\endinnercustomgeneric}
}

\newcustomtheorem{customthm}{Theorem}
\newcustomtheorem{customlemma}{Lemma}
\newcustomtheorem{customprop}{Proposition}
\newcustomtheorem{customdef}{Definition}

\newtheorem{prop}{Proposition}
\newtheorem{theorem}{Theorem}

\newcommand{\e}[2]{ $#1\!\pm\!#2$ }
\newcommand{\s}[2]{ #1\% (#2\%) }

\DeclareMathOperator\erf{erf}

\title{Worst Cases Policy Gradients}

\author{
  Yichuan Charlie Tang\\
  Apple Inc.\\
  \texttt{yichuan\_tang@apple.com}\\
  \And
  Jian Zhang\\
  Apple Inc.\\
  \texttt{\small jianz@apple.com}\\
  \And
  Ruslan Salakhutdinov\\
  Apple Inc.\\
  \texttt{\small rsalakhutdinov@apple.com}\\
}

\begin{document}
\maketitle

\begin{abstract}
Recent advances in deep reinforcement learning have demonstrated the capability of learning complex control policies from many types of environments. When learning policies for safety-critical applications, it is essential to be sensitive to risks and avoid catastrophic events. Towards this goal, we propose an actor-critic framework that models the uncertainty of the future and simultaneously learns a policy based on that uncertainty model. Specifically, given a distribution of the future return for any state and action, we optimize policies for varying levels of conditional Value-at-Risk. The learned policy can map the same state to different actions depending on the propensity for risk. We demonstrate the effectiveness of our approach in the domain of driving simulations, where we learn maneuvers in two scenarios. Our learned controller can dynamically select actions along a continuous axis, where safe and conservative behaviors are found at one end while riskier behaviors are found at the other. Finally, when testing with very different simulation parameters, our risk-averse policies generalize significantly better compared to other reinforcement learning approaches.
\end{abstract}
\vspace{-0.15in}

\section{Introduction}
\begin{wrapfigure}{r}{0.5\textwidth}
\vspace{-0.4in}
\includegraphics[width=0.48\textwidth]{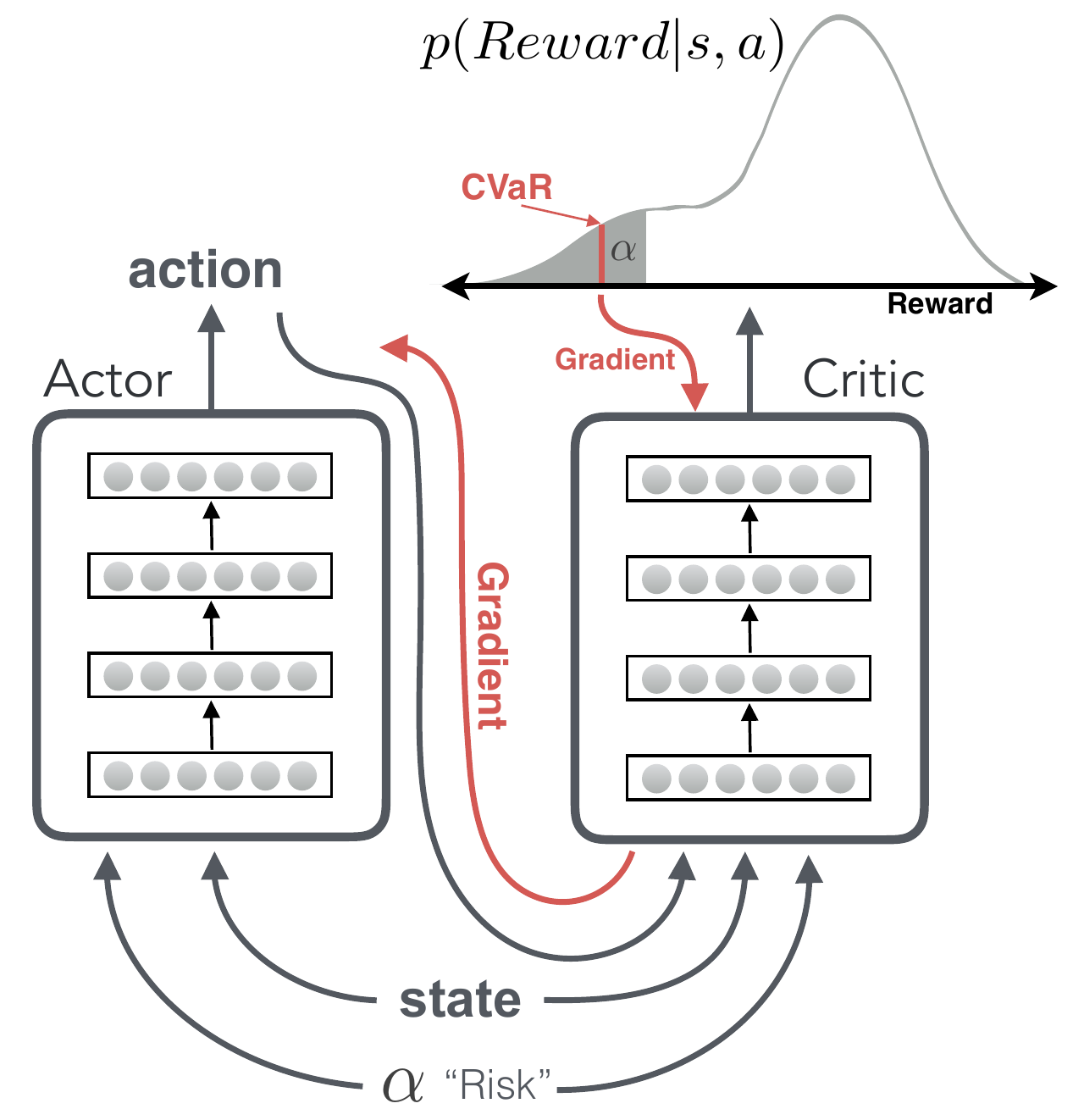}
\caption{ {Proposed architecture: the critic's estimated distribution over future returns is used to compute the conditional Value-at-Risk (CVaR) measure, which then generates gradients for learning the actor/policy network. See Sec.~\ref{sec:wcpg} for details.} }
\label{fig:wcpg_diags_new}
\vspace{-0.2in}
\end{wrapfigure}
One of the key challenges for building intelligent systems is developing the capability to make robust and safe sequential decisions in complex environments. Towards this goal, the recent breakthroughs in deep reinforcement learning (RL) are very encouraging and have led to super human performance in various video games and board games~\cite{mnih2015human,silver2017mastering,alphastarblog}.

As we move towards real-world applications and learning from increasingly diverse environments, we may encounter both \emph{parametric} and \emph{inherent} uncertainties due to the stochastic nature of the model and the environments~\cite{GarciaF15}. One cause of stochasticity is the inability to fully observe the state (e.g. intentions, beliefs) of other agents in a multi-agent environment. Robust handling of uncertainties and risks are a must before we can fully leverage the power of deep RL in real world safety-critical applications such as self-driving~\cite{shalev2017formal}. However, standard RL optimizes the average expected return and is not risk sensitive~\cite{SuttonB98}.

In this paper, we take a step towards this goal by developing a novel deep RL architecture that optimizes a risk-sensitive criterion. In contrast, the vast majority of existing deep RL techniques maximize the expected value over possible future returns~\cite{wang2016sample,LillicrapHPHETS15,schulman2017proximal,hessel2017rainbow}. Maximizing expectation, however, is not risk-sensitive as it does not explicitly penalize rare occurrences of catastrophic events. Working under the assumption that the future return is inherently stochastic, we first start by modeling its distribution. The risk of various actions can then be computed from this distribution. Specifically, we use the conditional Value-at-Risk~\cite{rockafellar2000optimization} as the criterion to maximize.
Our architecture is based on the actor-critic~\cite{KondaT00,LillicrapHPHETS15}, but we modify our critic to predict the full distribution over future returns instead of simply the expectation. 
The proposed framework (Fig.~\ref{fig:wcpg_diags_new}), which we call worst cases policy gradients (WCPG), learns a continuous family of policies, each optimizing for their respective level of risk: $\alpha \in [0.0, 1.0]$. Depending on the risk appetite, the learned policy can choose to act differently even from the same state.

After introducing our framework and the learning algorithm in Section~\ref{sec:wcpg}, we demonstrate the effectiveness of WCPG on two traffic simulation scenarios, where an agent must learn to safely interact with other agents and to achieve their own goals. In Section~\ref{sec:results}, we describe the details of our environments, training procedure, and quantitative results, where we find WCPG learns risk-averse policies which are significantly more robust in test scenarios.
\vspace{-0.1in}
\section{Preliminaries}\vspace{-0.1in}
We formulate our problem as a \emph{Markov Decision Process} (MDP) which consists of a state space $\mathcal{S}$ (a compact subset of $\mathbb{R}^d$), an action space $A = \mathbb{R}^m $, and a reward function $r(s,a): \mathcal{S} \times \mathcal{A} \rightarrow \mathbb{R}$. The model of the environment is $p(s'|s,a)$ which specifies the probability of transitioning to state $s'$ from state $s$ and executing action $a$. The policy function $\pi_\theta(a|s)$, parameterized by $\theta$, is the probability of choosing action $a$ given state $s$. $\rho^\pi(s)$ denotes the stationary distribution over the state space given that policy $\pi$ is followed. We denote the total discounted future return as $R^\gamma_t = \sum^\infty_{i=t} \gamma^{i-t} r(s_i,a_i)$, where $\gamma \in [0.0, 1.0]$ is the discount factor. The value function is $V^\pi(s) = \mathbb{E}[R_1^\gamma | S_1 = s, \pi]$ and the state-action value function is $Q^\pi(s, a) = \mathbb{E}[R_1^\gamma | S_1 = s, A_1=a,\pi]$. Reinforcement learning consists of a class of algorithms which can be used to find the optimal policy for MDPs~\cite{SuttonB98}.

Policy gradient methods directly optimize the policy parameters $\theta$ to maximize the expected total return.
Popular in continuous control, they compute the gradients with respect to the objective $J$: 
{ \small
\begin{equation}\label{eq:pg_objective}
J = \int_{S}\rho^{\pi}(s) \int_A \pi_\theta(a|s) Q^\pi(s,a) da \ ds
\end{equation} 
}%
Using the \emph{Policy Gradient Theorem} and the log-derivative trick~\cite{Williams92,SuttonMSM00}, the gradient of the objective can be written as:\vspace{-0.05in}
{ \small
\begin{equation}
\label{eq:pgt}
\nabla_\theta J = \mathbb{E}[\nabla_\theta \log \pi_\theta(a|s) Q^\pi(s,a)],
\end{equation}
}%
where the gradient does not dependent on the state distribution $\rho^\pi(s)$. \cite{SchulmanMLJA15} have shown that it may be beneficial to substitute $Q^\pi(s,a)$ with other terms, where it can be one of several expressions: $Q^\pi(s, a)$, $V^\pi(s)$, or the advantage function $A^\pi(s, a) =  Q^\pi(s, a)-V^\pi(s)$, leading to a lower variance of the gradients. 
\vspace{-0.1in}
\subsection{Deep Deterministic Policy Gradients} \vspace{-0.1in}
\label{sec:ddpg}
While Eq.~\ref{eq:pgt} holds for any stochastic policy,~\cite{SilverLHDWR14} introduced the Deterministic Policy Gradient (DPG) theorem for deterministic policies, whose gradients can be estimated more efficiently. DPG states (under certain regularity conditions) that for a \emph{deterministic} policy $\pi$:
\vspace{-0.05in}
\begin{equation}\label{eq:dpg}
\nabla_\theta J(\pi_\theta) = \mathbb{E}_{s \sim \rho}[\nabla_\theta \pi_\theta(s) \nabla_a Q^\pi(s,a)|_{a=\pi_\theta(s)}].
\end{equation}
The deep deterministic policy gradient (DDPG) framework~\cite{LillicrapHPHETS15} extended DPG to large continuous state action space environments. DDPG works well for continuous control and is a popular variant of the \emph{actor-critic}~\cite{SuttonMSM00,KondaT00} policy gradients method. A critic function learn to estimate $Q^\pi(s,a)$ using temporal difference bootstrapping and provides the training signal for the actor (policy network). Variants of DDPG have also been explored, including actors with parametrized actions~\cite{HausknechtS15}.

\section{Worst Cases Policy Gradients}\vspace{-0.1in}
\label{sec:wcpg}
Standard reinforcement learning maximizes for the expected (possibly discounted) future return~\cite{SuttonB98}. However, maximizing for average return is not sensitive to the possible risks when the future return is stochastic, due to the inherent randomness of the environment not captured by the observable state. For example, when the return distribution has high variance or is heavy-tailed, finding a policy which maximizes the expectation of the distribution might not be ideal: a high variance policy (and therefore higher risk) that has higher return in expectation is preferred over low variance policies with lower expected returns. Instead, we want to learn more robust policies by minimizing long-tail risks, reducing the likelihoods of bad outcomes. 

\begin{figure}[t]
\begin{center}
\includegraphics[width=0.7\linewidth]{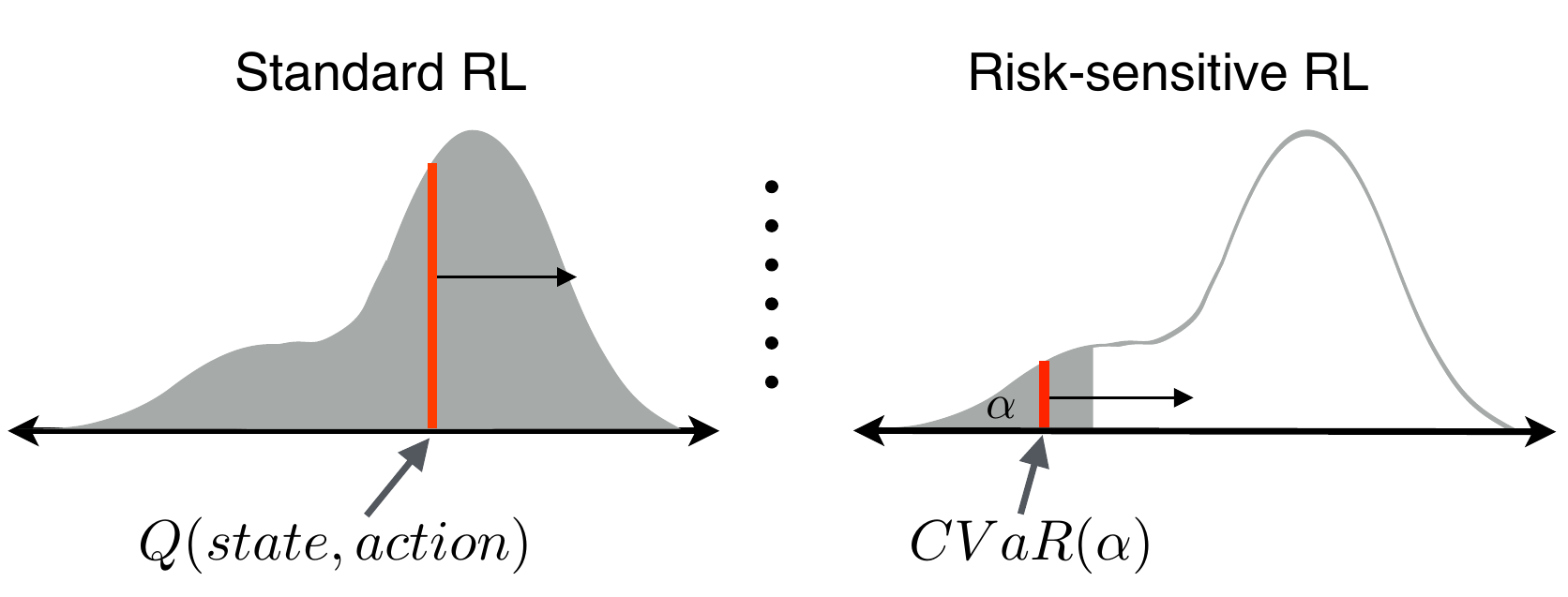}
\end{center}
\vspace{-0.15in}
\caption{{\textbf{Left}: standard RL maximizes for the expected future return (red line). \textbf{Right}: WCPG optimizes the risk-sensitive $CVaR(\alpha)$, which is the expected value up to the $\alpha$-percentile of the distribution of future return. } }
\label{fig:cvar_diag}
\vspace{-0.3in}
\end{figure}
We start by first modeling the distribution of expected future returns for all joint states and actions. For a single particular state-action pair $\{s,a\}$, let $R$ be the random variable for future return, and $ p^\pi(R|s,a) $ be the probability \emph{distribution} of $R$ given our current state and action while following policy $\pi$. We can recover the widely used state-action value function: $Q^\pi( s, a) = \mathbb{E}_{p^\pi} [ R| s, a]$. In the illustration of Fig.~\ref{fig:cvar_diag}, the x-axis is $R$ and the y-axis is the probability density. In the left panel, the red line marks the expected value of the distribution or $Q^\pi( s, a)$. Typically, we optimize the policy by changing the shape of the density so as to increase $Q^\pi( s, a)$ or shift the red line to the right. However, for risk-averse learning, it is desirable to maximize the expected \emph{worst} cases performance instead of the average-case performance. 
We can accomplish this by optimizing for the conditional Value-at-Risk (CVaR) criterion~\cite{rockafellar2000optimization} instead. Intuitively, CVaR represents the expected return should we experience the bottom $\alpha$-percentile of the possible outcomes. 
The right panel of Fig.~\ref{fig:cvar_diag} demonstrates that by optimizing for CVaR, we seek policies that will shift the tail-end of the distribution to the right, thus reducing the probability of experiencing a very negative event.

Formally, the criterion that we care about is the $\alpha$-percentile of the distribution over returns. This is captured by the conditional Value-at-Risk:\vspace{-0.1in}
\begin{equation}
CVaR_\alpha \doteq \mathbb{E}_{p^\pi}[R|R \leq pcntl(\alpha)],
\label{eq:cvar}
\end{equation}
where $ pcntl(\alpha) $ is the $\alpha$-percentile of $p^\pi$. As $\alpha \rightarrow 0$, the policy will focus on performing well in the ``\emph{worst cases}" scenarios. 
Computing CVaR metric directly for long time horizons (e.g. a brute-forced approach like sampling~\cite{Tamar14}) would be prohibitively expensive. As an alternative, we extend DDPG (an actor-critic architecture) so that the critic learns to model the distribution over expected total discount reward. Provided a good distributional critic can be learned, CVaR can be computed from the distribution and we can train the actor by backpropagating the gradient back through the actor or policy network.

At this point, the reader might wonder if it is possible to first train (offline) a critic for $p^\pi(R| s,a ) $ and then obtain a risk-sensitive policy (online) simply by selecting actions to maximize the $\alpha$-percentile of $p^\pi(R| s,a )$ at every state. However, this approach would fail as $p^\pi(R| s,a )$ is conditioned on executing policy $\pi$ for all subsequent steps. Selecting the action (which modifies the original policy) that gave the best $\alpha$ percentile at time $t$ is not meaningful if we revert to policy $\pi$ for time $t+1$ and beyond. In other words, multiple actions at a given state $s$ would have similar distributions over $R$, as a single action (especially continuous) at time $t$ can be nullified by future actions.
\vspace{-0.1in}
\subsection{Distributional Critic}\vspace{-0.1in}
To learn the critic, we must define the equivalent Bellman operator for distributions. Let us first define $Z(s,a) \doteq p^\pi(R| s,a)$ as the distribution over total future (possibly discounted) return generated by executing policy $\pi$ until reaching a terminal state. We define $P^\pi$ as the transition operator:
\vspace{-0.05in}
\begin{align}
P^\pi Z(s,a) \doteq Z(s',a'), \ s' \sim p(\cdot|s,a), \  a' \sim \pi(\cdot|s'),
\vspace{-0.05in}
\end{align}
and the distributional Bellman operator~\cite{Sobel82,MorimuraSKHT10,TamarCM16} $\mathcal{T}^\pi$ is:
$
\mathcal{T}^\pi Z(s,a) \doteq r(s,a) + \gamma P^\pi Z(s,a).
$
Here, we approximate $Z(s,a)$ up to its 2nd-order statistics, the mean and the variance. Defining the mean as $Q^\pi(s,a) = \mathbb{E}_{p^\pi} [ R| s, a] $ and the variance $\Upsilon^\pi(s,a) = \mathbb{E}_{p^\pi} [ R^2| s, a] - (Q^\pi(s,a))^2 $, we have $Z^\pi(s,a) \sim \mathcal{N}( Q^\pi(s,a), \Upsilon^\pi(s,a)  ) $. Modeling $Z$ as a Gaussian leads to a closed-form calculation of the CVaR, as we will see shortly\footnote{This is critical as we must compute CVaR for every update step, for every state-action tuple, and for different risk $\alpha$ parameters. Sampling to estimate CVaR would be prohibitively expensive computationally.}.

There are two sets of projection equations for the distributional Bellman operator. The projection for $Q^\pi$ (mean) is the standard Bellman equation, similar as in Q-learning: $Q^\pi(s,a) = r(s,a) + \gamma \sum_{s'} p(s'|s,a) Q^\pi(s',a') $. The projection for $\Upsilon^\pi$ (variance) is:
\begin{prop} The following equation for the variance of the future return exists.
\begin{align}\label{eq:dis_bell}
\vspace{-0.05in}
\Upsilon^\pi(s,a) =& r(s,a)^2 + 2 \gamma r(s,a)\sum_{s'} p(s'|s,a) Q^\pi(s',a')  \nonumber \\ 
&+ \gamma^2 \sum_{s'} p(s'|s,a) \Upsilon^\pi(s',a') + \gamma^2 \sum_{s'} p(s'|s,a) Q^\pi(s',a')^2 - Q^\pi(s,a)^2.
\vspace{-0.05in}
\end{align}
\end{prop}
\vspace{-0.15in}
A straight-forward proof is given in the Appendix. Eq.~\ref{eq:dis_bell} is the parametric distributional Bellman operator for the distributional state-action value function. In practice, due to a relatively large continuous state and action spaces, we use function approximation for representing the critic. Specifically, a neural network parameterized by $\omega$ is used for estimating the critic's mean and variance:
$f_{critic}(s,a|\omega) \rightarrow \lbrace \hat{Q}^\pi(s,a), \hat{\Upsilon}^\pi(s,a) \rbrace $.

We learn the critic by using the Temporal Difference (TD) learning, where the TD error is obtained by assuming Gaussianity (maximum entropy distribution) and using the Wasserstein metric as the loss. 
While other losses such as KL-divergence could also be used, the distributional Bellman operator using the $p$-th Wasserstein metric has been shown to be a contraction operator for policy evaluations~\cite{BellemareDM17}. The $p$-th Wasserstein distance between two probability distributions $u$ and $v$ is defined as~\cite{olkin1982distance}:
\vspace{-0.05in}
\begin{equation}
W_p(u, v) \doteq \Big ( \int_0^1 \big | F^{-1}_u(s) - F^{-1}_v(s) \big |^p \ ds  \Big )^{1/p},
\end{equation}
where $F^{-1}$ is the inverse cumulative distribution function (CDF). Assuming $u\sim \mathcal{N}(\mu_1,C_1)$ and $v\sim \mathcal{N}(\mu_2,C_2)$, the 2-Wasserstein distance simplifies to:
\vspace{-0.05in}
\begin{equation}
W_2(u,v) = || \mu_1 - \mu_2 ||^2_2 + trace(C_1 + C_2-2(C_2^{1/2}C_1C_2^{1/2})^{1/2})
\label{eq:wasserstein}
\end{equation}
The critic will try to minimize the Wasserstein distance by backpropagating the gradient $\frac{\partial W_2}{\partial \omega}$.
\subsection{CVaR Actor}\vspace{-0.1in}
For every state action $\{s,a\}$, we can use the critic's estimated $\hat{Q}^\pi(s,a), \hat{\Upsilon}^\pi(s,a)$ (mean and variance of a Gaussian) to compute the $CVaR_\alpha$ measure in closed-form:
\vspace{-0.05in}
\begin{align}\label{eq:actor_loss}
\Gamma^\pi(s,a,\alpha) \doteq CVaR_\alpha = Q^\pi(s,a) -  (\phi(\alpha) \slash \Phi(\alpha)) \sqrt{ \Upsilon^\pi(s,a) },
\end{align}
where $\phi(\cdot)$ is the standard normal distribution and $\Phi(\cdot)$ is its CDF:
$
\Phi(x) = \frac{1}{2} \big ( 1+ \erf (x/ \sqrt{2}) \big )
$. We use $\Gamma^\pi(s,a,\alpha)$ to explicitly denote the reliance of $CVaR$ on both the state, action, as well as a specific policy $\pi$. We stress that $\Gamma^\pi(s,a,\alpha)$ is a scalar term computable at every state-action pair $\{s,a\}$. In plain language, $\Gamma^\pi(s,a,\alpha)$ is the expected future CVaR when in state $s$ and executing action $a$, and following policy $\pi$ hereafter. 

We now replace the standard objective of Eq.~\ref{eq:pg_objective} by the risk-averse $CVaR_\alpha$ objective:
\begin{equation}
J_\alpha = \int_S \big [ \rho^\pi(s) \int_A \pi_\theta(a|s) \Gamma^\pi(s,a,\alpha) \ da  \big ] ds.
\end{equation}
\begin{theorem}
Suppose that the MDP satisfied conditions A.1 and A.2 (see Appendix), then:
\begin{align}
\nabla_\theta J_\alpha = \mathbb{E}_{s\sim\rho, a\sim\pi} [ \nabla_\theta \log \pi_\theta(a|s) \Gamma^\pi(s,a,\alpha) ]. 
\label{eq:wc_pg}
\end{align}
Proof. The proof mainly follows the original Policy Gradient Theorem (PGT)~\cite{SuttonMSM00}, see Appendix.
\end{theorem}

Given Eq.~\ref{eq:wc_pg}, we follow DPG from Eq.~\ref{eq:dpg} to derive the equivalent deterministic policy gradient, where we backpropagate the CVaR loss gradients through the critic and down to the actor/policy networks. We obtain the deterministic gradient for the actor network by following a similar derivation as in~\cite{SilverLHDWR14}:
\begin{small}
\begin{align}
\nabla_\theta J_\alpha &= \mathbb{E}_{s\sim\rho, \alpha} [ \nabla_\theta \pi(a|s,\alpha) \nabla_a \Gamma^\pi(s,a,\alpha) ] \nonumber  \\
&= \mathbb{E}_{s\sim\rho, \alpha} \big [ \nabla_\theta \pi(a|s,\alpha) \nabla_a \hat{Q}^\pi(s,a,\alpha)
	- (\phi(\alpha) \slash \Phi(\alpha)) \nabla_\theta \pi(a|s,\alpha) \nabla_a \sqrt{ \hat{\Upsilon}^\pi(s,a,\alpha)} \big ].
\label{eq:dwcpg}
\end{align}
\end{small}%
Note that our new objective $J$ is dependent on ``risk level" $\alpha$, and the question now arises as to how to choose the percentile $\alpha$, which ranges from $0.0$ to $1.0$. One strategy is to discretize $\alpha$ into $N$ discrete values and train $N$ separate policy networks optimizing for $N$ different settings. Instead of this naive approach with $N$ times more parameters, we learn a single network which is conditioned on $\alpha$ as an additional input (see Fig.~\ref{fig:wcpg_diags_new}). The advantage of this approach is that we can learn a continuous family of $\alpha$ conditional policies $\pi_\theta(a|s,\alpha)$ with varying risk-sensitivity.

During training, a different input $\alpha$ will have a different loss function $J_\alpha$. To train for all $\alpha$s, we uniformly sample $\alpha\sim Uniform(0,1)$ during the \emph{start} of an episode and fix $\alpha$ for the entirety of that episode. During inference, $\pi$ 
can output \emph{different} actions given the same exact state $s$, conditioned on the setting of $\alpha$. Intuitively, a small $\alpha$ leads to conservative actions while a larger $\alpha$ leads to more aggressive actions.

We now have the learning equations for both the actor and the critic. We employ an off-policy training algorithm by using an experience replay buffer. Algorithm~\ref{alg:wc_pg} in the Appendix outlines the entire WCPG training procedure. Fig.~\ref{fig:wcpg_diags_new} provides an overall illustration of the entire WCPG architecture and gradient flow during training.

\begin{figure}[t]
\begin{minipage}[c]{0.5\textwidth}
  \includegraphics[width=1\textwidth]{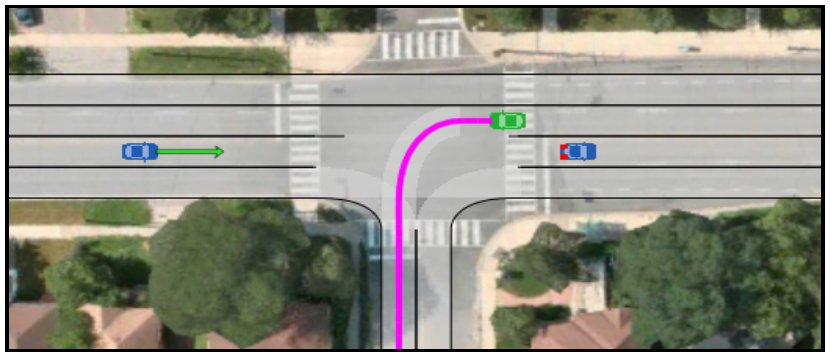}
  \captionof{figure}{{Unprotected left turn environment.}}
  \label{fig:unp_left_env}
\end{minipage}
\begin{minipage}[c]{0.5\textwidth}
  \includegraphics[width=1\textwidth]{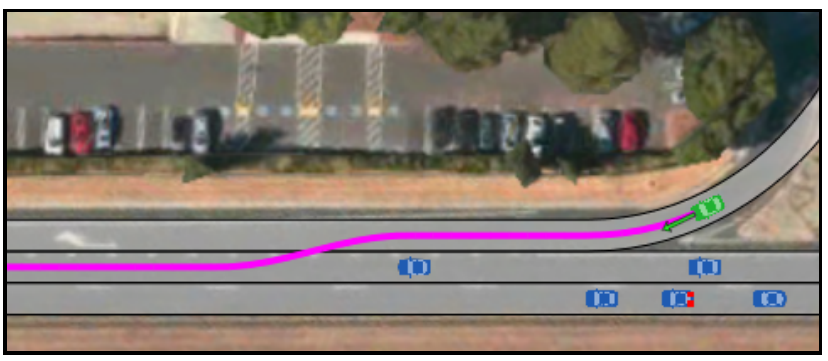}
  \captionof{figure}{{Highway merge environment.}}
  \label{fig:highway_merge_env}
\end{minipage}
\vspace{-0.2in}
\end{figure}

\vspace{-0.1in}
\section{Experimental Results}
\label{sec:results}
\vspace{-0.1in}
\subsection{Environments}\vspace{-0.1in}
\label{sec:environments}
We tested our algorithm on two continuous-action 2D driving environments, focusing on two of the more critical scenarios: unprotected turns and merges (Figs~\ref{fig:unp_left_env},~\ref{fig:highway_merge_env}). In our environments, each agent is a vehicle, with the goal of getting from point A to point B while staying on the road and avoiding collisions. We simulate the vehicle dynamics using a discrete time kinematics bicycle model~\cite{KongPSB15}. The simulation timestep is 100 milliseconds. We restrict our action space to be the acceleration of the ego vehicle\footnote{\emph{Ego} refers to the vehicle for which we are learning a policy. We bound the acceleration and deceleration of all vehicles to $\pm4 \ m/s^2$, similar to that of a typical real vehicle.}. 
The steering is determined by the Stanley controller, a non-linear closed loop feedback steering controller~\cite{thrun2006stanley}. The physical dimension of our simulation environments is approximately 200 meters by 200 meters, where agents are randomly spawned on specific ``birth" lanes with a given probability. Ego's initial velocity is randomly chosen between 5 to 20 m/s.

As a part of our multi-agent environments, non-ego agents are controlled by rule-based behaviors. Their velocity is randomly chosen and they have the ability to perform adaptive cruise control: slowing down and speeding up according to the vehicle in front of them. They can also perform safe lane changes by dynamically planning a smooth trajectory in order to merge from one lane to another. To introduce more realistic behaviors, the non-ego agents, during spawning, each samples randomly from one of three behaviors (when close to ego): \emph{yield, ignore}, or \emph{accelerate}, mimicking human drivers who might be conservative, distracted, or aggressive.

The reward for catastrophic failure (e.g. collision) is $-50.0$, while the reward for successful completion of a maneuver depends on time-to-completion: $50 \times e^{-steps/50}+10$. A failure to complete the maneuver results in a reward of $0$. We initially attempted to learn safer policies by simply scaling up the collision penalties (e.g. $-1000.0$), however we found that the resulting policies learned to be overly conservative and often did not even attempt the maneuvers. See Appendix for more details.

\vspace{-0.1in}
\subsection{WCPG Network}\vspace{-0.1in}
Our network is an actor-critic based on DDPG~\cite{LillicrapHPHETS15} with two modifications. The first is that an additional~$\alpha$ CVaR input is added to both the actor and the critic. The second is that instead of producing a scalar value for approximating $Q(s,a)$, the critic outputs parameters which govern the distribution of future returns. We use the \emph{softplus}\footnote{$f(x) = \log (1 + \exp(x))$} function to guarantee that the critic's estimation of variance will always be positive. See Fig.~\ref{fig:wcpg_diags_new} for the diagram of our network model. The input to our network is $16$ dimensional and consists of encoding the closest vehicles around ego (in sorted order) and their states: $(x,y)$ position, heading, and velocity. The actor consists of 3 hidden ReLU layers with 32 hidden units each. The output of the actor is ego's acceleration. The critic network consists of 4 hidden layers with 64 hidden units each.  We did not find the performance to be particularly sensitive to the choice of number of layers and layer size. See Appendix for more details.

Off-policy training is performed with an experience replay buffer of up to $10^6$ $\lbrace s,a,r,s' \rbrace$ tuples. The learning rate for both the actor and the critic is $0.0001$, and the minibatch size is $512$. The same action is executed $4$ times in a row for a total of $400$ ms. Exploration noise is Gaussian with a standard deviation of 2.0. Input mean and variance for normalization is calculated in an online moving average fashion. $\alpha$ is randomly sampled from $[0.01,1.0]$. Training is run until convergence for 5000 episodes, where each episode can last up to 30 seconds in simulation (300 timesteps).

\vspace{-0.1in}
\subsection{Testing Performance}
\begin{figure*}
    \begin{subfigure}[t]{0.36\textwidth}
        \includegraphics[width=\textwidth]{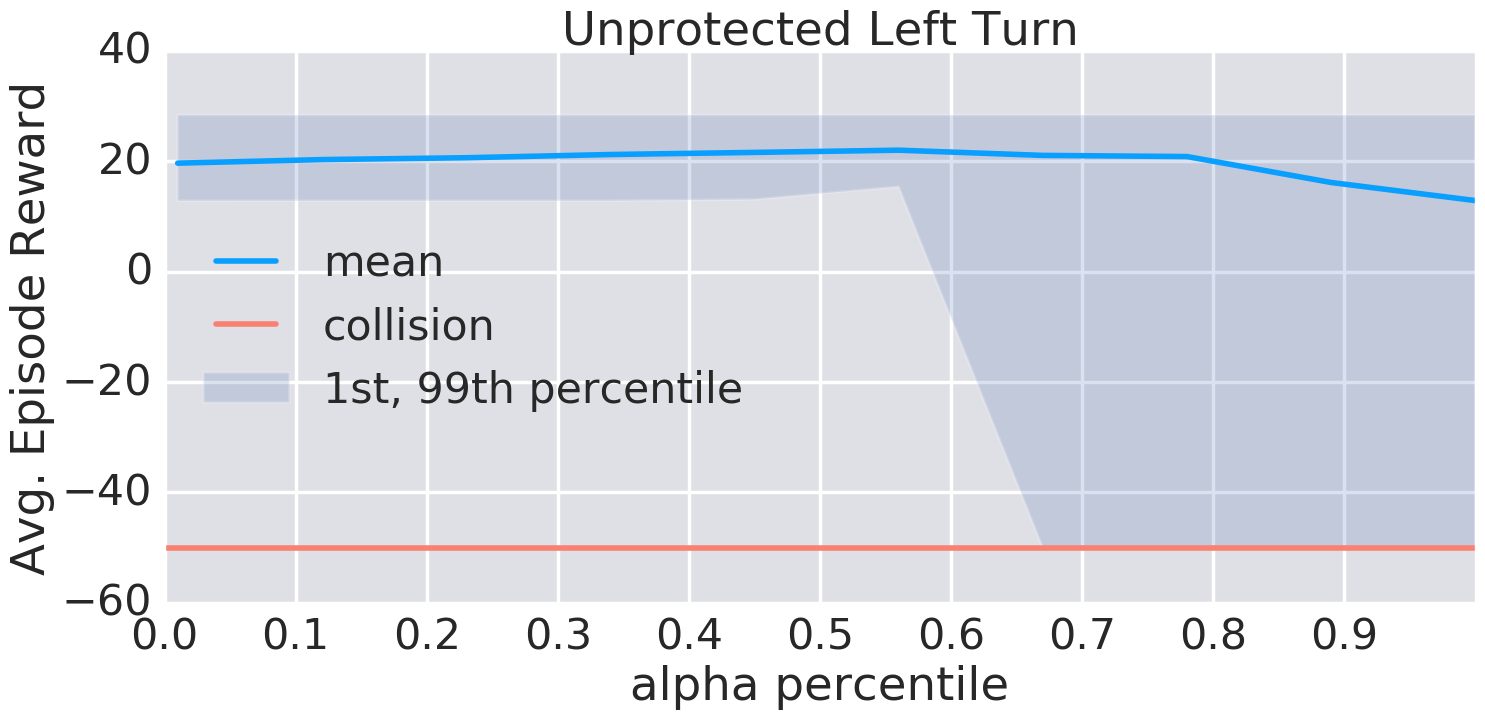}        
        \label{fig:unp_left_alpha_plots}
    \end{subfigure}
    \begin{subfigure}[t]{0.36\textwidth}
        \includegraphics[width=\textwidth]{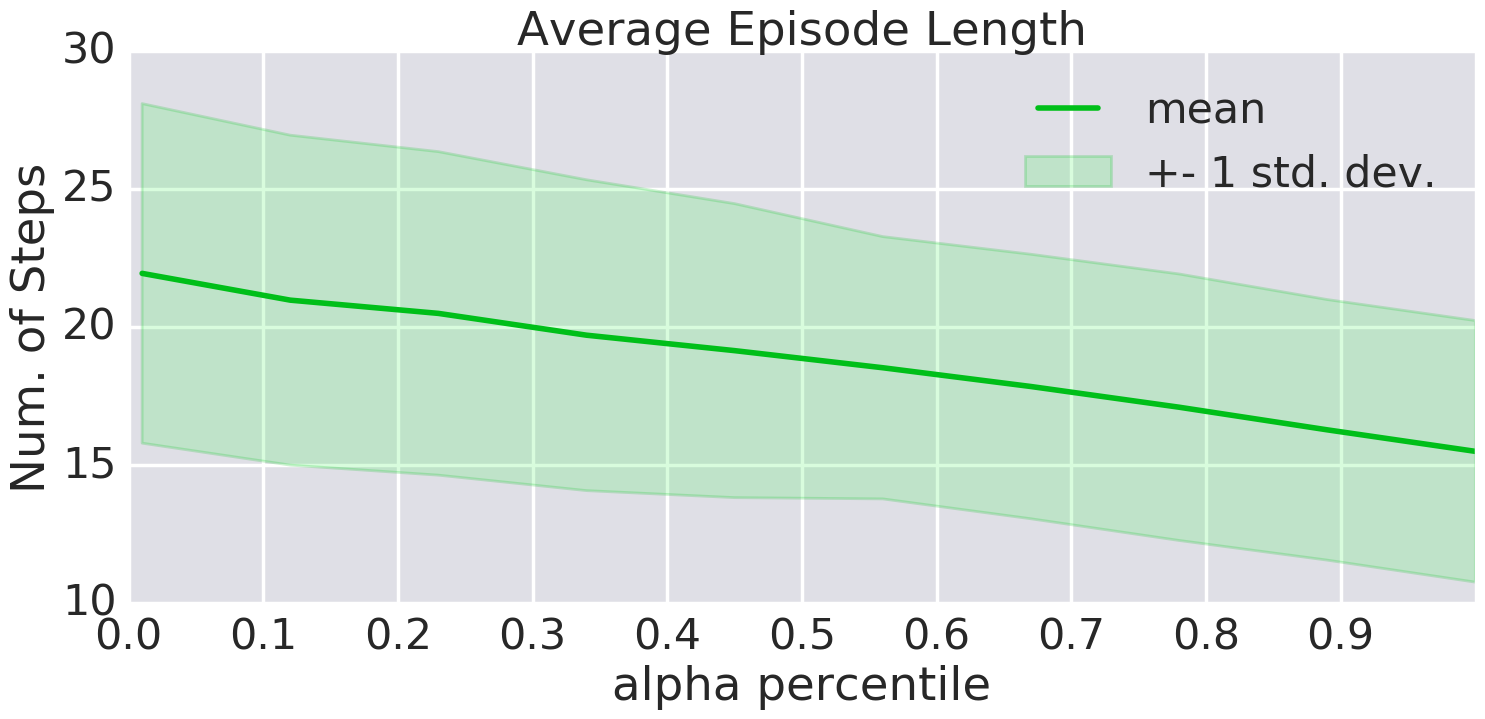}
    \end{subfigure}
    \begin{subfigure}[t]{0.27\textwidth}
        \includegraphics[width=\textwidth]{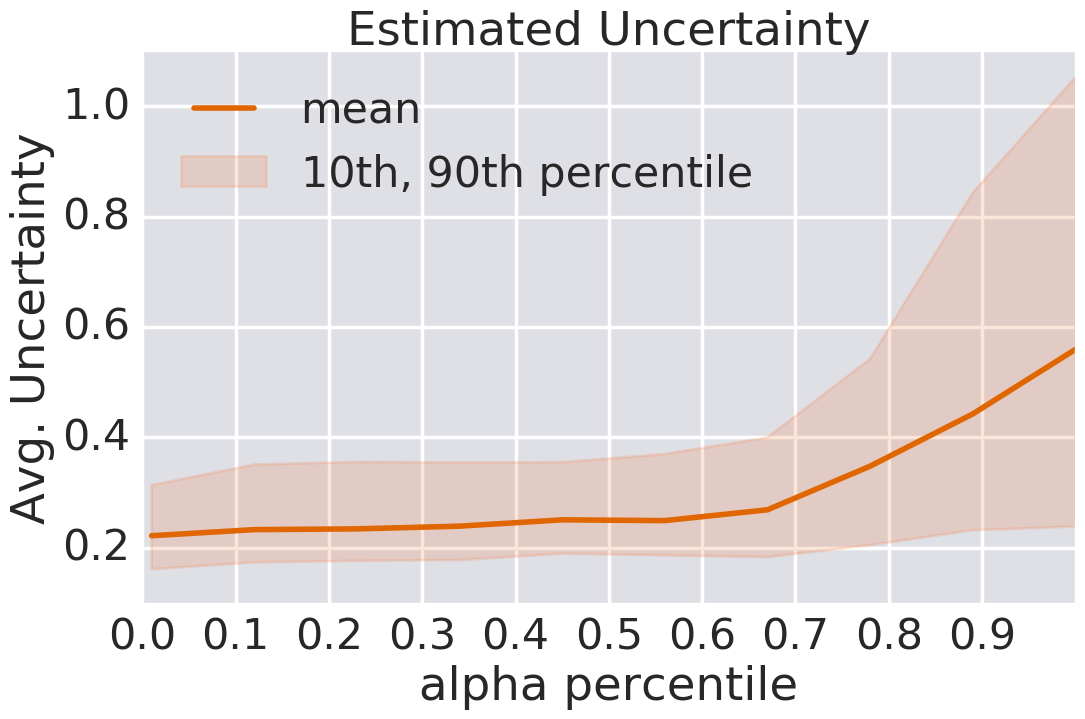}
    \end{subfigure}
    \vspace{-0.1in}
    \\
\hspace{-0.2in}
    \begin{subfigure}[t]{0.36\textwidth}
        \includegraphics[width=\textwidth]{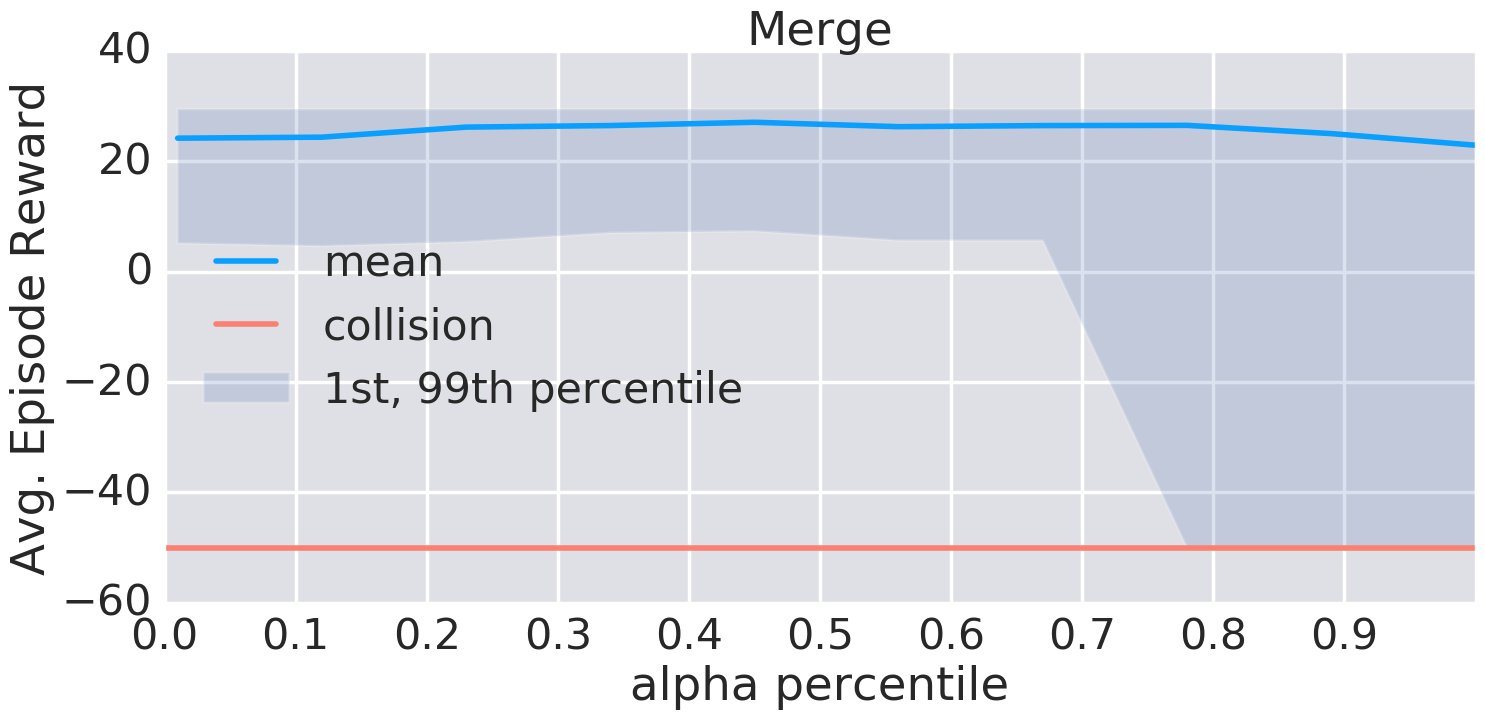}
         \label{fig:merge_alpha_plots}
    \end{subfigure}
    \begin{subfigure}[t]{0.36\textwidth}
        \includegraphics[width=\textwidth]{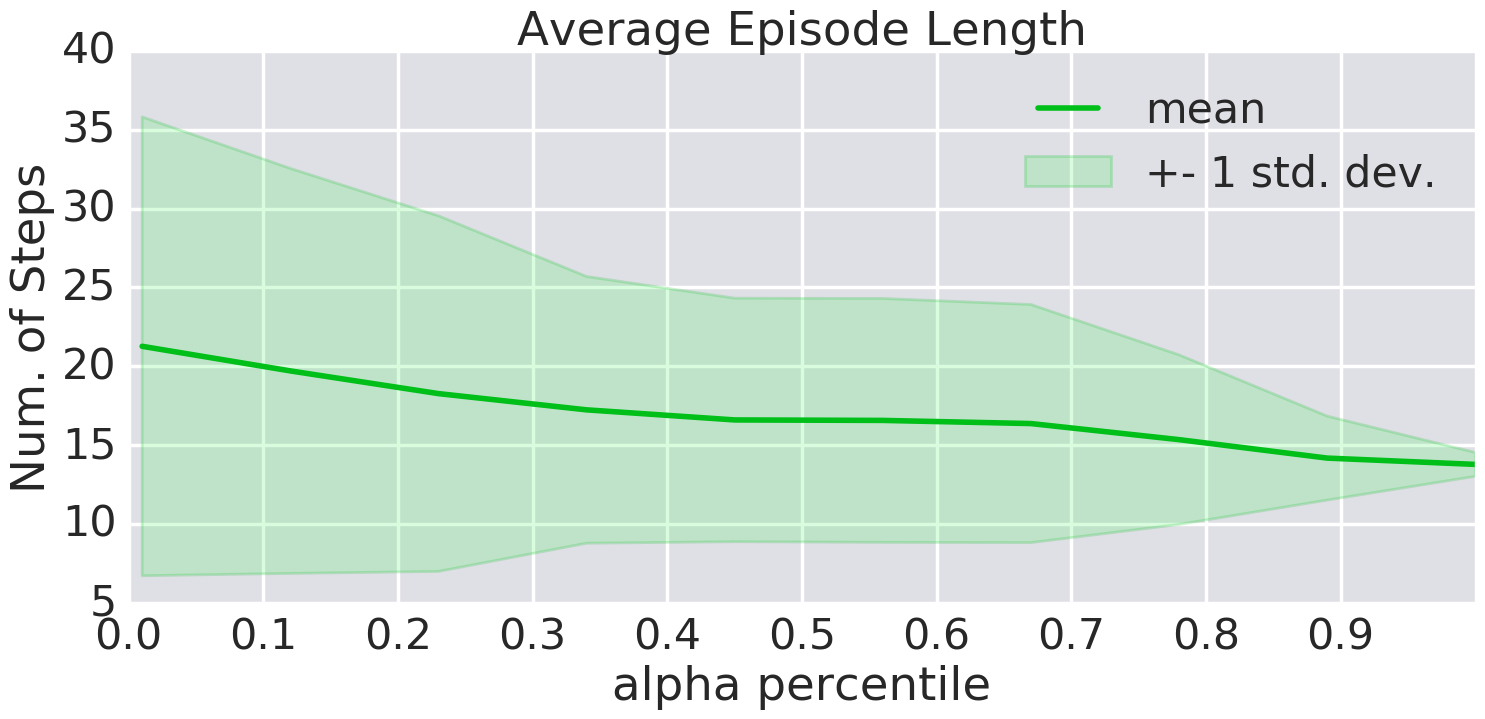}
    \end{subfigure}
    \begin{subfigure}[t]{0.27\textwidth}
        \includegraphics[width=\textwidth]{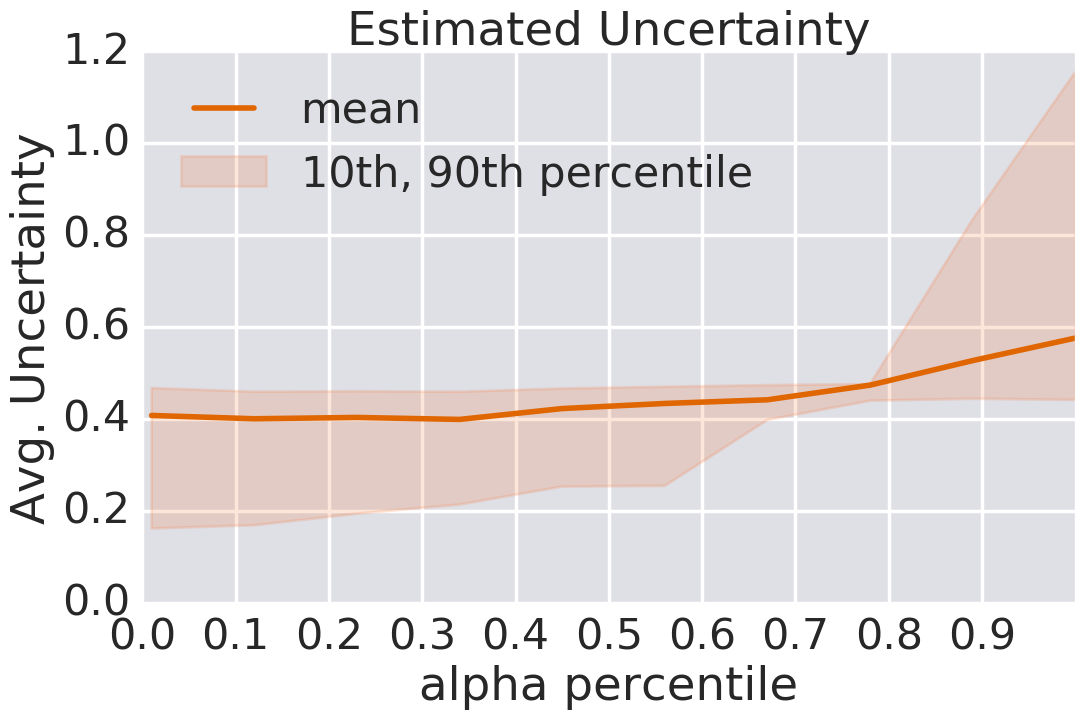}
    \end{subfigure}
    \vspace{-0.1in}
    \caption{\small {Testing performance for the unprotected turn (top row) and merge (bottom row) environments. \textbf{Left}: we plot the distribution of episodic reward and their 1st and 99th percentile as a function of $\alpha$. Red line indicates the reward incurred for a collision. The policy function successfully learned to be more risk-averse as $\alpha \rightarrow 0$. \textbf{Middle}: the number of steps to complete an episode increases as $\alpha$ decreases, due to the more conservative behavior of the learned policy. \textbf{Right}: average episodic standard deviation of future return estimated by the critic, as a function of $\alpha$.}}
    \label{fig:results_plots}
    \vspace{-0.1in}
\end{figure*}

\vspace{-0.1in}
We now look at WCPG performs on test environments\footnote{The random seeds for the testing environments are different compared to training environments.}. Results are shown for the two environments in Fig.~\ref{fig:results_plots}. In the left panel, we can see that as $\alpha$ decreases, the policy becomes risk-averse and more robust, reducing the probability of collisions. In the middle, we can see that smaller $\alpha$s leads to more conservative behavior (e.g. waiting for a wider gap), increasing the time-to-completion for both environments. Finally, the right panel shows that the critic's own estimate of uncertainty grows with an increasing $\alpha$. This also means that the actor also has a higher risk-tolerance with an increasing $\alpha$.

\begin{figure}[t]
\begin{minipage}[t]{1.0\textwidth}
\scriptsize
  \setlength{\tabcolsep}{2pt} 
  \captionof{table}{{ Left turn environment (extrapolation) - Collision rate $\%$ (Success rate $\%$) }} \vspace{-0.1in}
\makebox[1\textwidth][c]{
    \begin{tabular}{rc >{\columncolor{lightgray2}}l >{\columncolor{lightgray2}}l >{\columncolor{lightgray2}}l >{\columncolor{lightgray2}}l >{\columncolor{lightgray2}}l l l>{\columncolor{blue}}l >{\columncolor{blue}}l }
\toprule
 \multicolumn{2}{c}{Params}  & \multicolumn{5}{c}{WCPG (varying $\alpha$)} & \multicolumn{4}{c}{State-of-the-art baselines} \\
\cmidrule(lr){1-2}  \cmidrule(lr){3-7} \cmidrule(lr){8-11}
vel. & spwn. &   0.02  & 0.1 & 0.3  & 0.6 & 1.0 & DDPG & PPO & C51 & D4PG \\ 
\midrule
(train)& 1\%&	\s{0}{100}  & \s{0}{100}  & \s{0}{100}   & \s{2}{98}  & \s{15}{85} & \s{0}{100} & \s{4}{96} 	 & \s{2}{98} & \s{2}{98} \\
\midrule
+5 m/s&   5\%&	\s{0}{100}  & \s{0}{100}  & \s{0}{100}   & \s{0}{100} & \s{15}{85} & \s{13}{87} & \s{24}{76} & \s{10}{90} & \s{4}{96} \\
+10 m/s&  5\%&	\s{0}{100}  & \s{0}{100}  & \s{0}{100}   & \s{0}{100} & \s{15}{85} & \s{18}{82} & \s{40}{60} & \s{11}{89} & \s{4}{96} \\
+15 m/s&  5\%& 	\s{0}{100}  & \s{0}{100}  & \s{0}{100}   & \s{3}{97}  & \s{21}{79} & \s{20}{80} & \s{49}{51} & \s{14}{86} & \s{2}{98} \\
+0 m/s&   2\%&	\s{0}{100}  & \s{0}{100}  & \s{0}{100}   & \s{0}{100} & \s{19}{81} & \s{7}{93} &  \s{15}{85} & \s{4}{96} &  \s{5}{95} \\
+0 m/s&   8\%& 	\s{0}{100}  & \s{0}{100}  & \s{0}{100}   & \s{3}{97}  & \s{26}{74} & \s{6}{94} &  \s{15}{85} & \s{5}{95} &  \s{1}{99} \\
+10 m/s&  8\%& 	\s{0}{100}  & \s{0}{100}  & \s{0}{100}   & \s{2}{98}  & \s{23}{77} & \s{19}{81} & \s{40}{60} & \s{11}{89} & \s{2}{98} \\
\bottomrule
\end{tabular}%
}
\label{tab:generalization1}
\vspace{0.1in}
\end{minipage}
\begin{minipage}[t]{1.0\textwidth}
\scriptsize
  \setlength{\tabcolsep}{2pt} 
  \captionof{table}{{ Merge environment (extrapolation) - Collision rate $\%$ (Success rate $\%$) } } \vspace{-0.1in}
\makebox[1\textwidth][c]{
    \begin{tabular}{rc >{\columncolor{lightgray2}}l >{\columncolor{lightgray2}}l >{\columncolor{lightgray2}}l >{\columncolor{lightgray2}}l >{\columncolor{lightgray2}}l l l>{\columncolor{blue}}l >{\columncolor{blue}}l }
\toprule
 \multicolumn{2}{c}{Params}  & \multicolumn{5}{c}{WCPG (varying $\alpha$)} & \multicolumn{4}{c}{State-of-the-art baselines} \\
\cmidrule(lr){1-2}  \cmidrule(lr){3-7} \cmidrule(lr){8-11}
vel. & spwn. & 0.02 & 0.1 & 0.3  & 0.6 & 1.0 & DDPG & PPO & C51 & D4PG \\ 
\midrule
(train)& 1\%&	\s{0}{100}  & \s{0}{100}& \s{0}{100}& \s{0}{100}& \s{8}{92}  & \s{0}{100} & \s{0}{96} 	 & \s{0}{100} & \s{6}{94} \\
\midrule
+5 m/s&   1\%&	\s{0}{4}  & \s{1}{5}  & \s{0}{54}   & \s{2}{90} & \s{10}{88} & \s{4}{96} & \s{31}{68} & \s{9}{91} & \s{9}{91} \\
+10 m/s&  1\%&	\s{0}{2}  & \s{0}{5}  & \s{0}{50}   & \s{0}{93} & \s{8}{92} & \s{3}{97} & \s{26}{73} & \s{5}{95} & \s{8}{92} \\
+15 m/s&  1\%& 	\s{0}{1}  & \s{0}{5}  & \s{0}{57}   & \s{1}{94} & \s{9}{91} & \s{1}{99} & \s{24}{76} & \s{6}{94} & \s{9}{91} \\
+0 m/s&   2\%&	\s{1}{7}  & \s{0}{11} & \s{1}{50}   & \s{2}{78} & \s{18}{76} & \s{5}{95} & \s{40}{59} & \s{8}{92} & \s{14}{86} \\
+0 m/s&   3\%& 	\s{0}{6}  & \s{0}{10} & \s{0}{43}   & \s{2}{70} & \s{17}{79} & \s{5}{95} & \s{39}{59} & \s{9}{91} & \s{12}{88} \\
+10 m/s&  3\%& 	\s{0}{4}  & \s{0}{8}  & \s{1}{43}   & \s{2}{76} & \s{11}{79} & \s{4}{96} & \s{33}{67} & \s{5}{95} & \s{10}{90} \\
\bottomrule
\end{tabular}%
}
\label{tab:generalization2}
\end{minipage}
\end{figure}

\vspace{-0.1in}
\subsection{Extrapolation Performance}\vspace{-0.1in}
A challenging test is in the ability for policies to extrapolate, where the parameters of testing environments are ``out-of-distribution" or extrapolated from the training distribution. Specifically, we significantly increase the upper-bound of non-ego agents' velocity by an additional
$15$ m/s and the spawn rate upwards of $8\%$. The total number of spawned agents is also increased (up to 4 additional agents) to make the scene denser.

Tables~\ref{tab:generalization1},~\ref{tab:generalization2} show the performance of WCPG and baselines on the extrapolation environments. We report collision rates for 100 random trials. Our results are highlighted in green. The training results are reported in the first row of each environment. We compare with 4 state-of-the-art algorithms in DDPG, proximal policy optimization (PPO)~\cite{schulman2017proximal}, C51/Rainbow~\cite{BellemareDM17,hessel2017rainbow}, and D4PG~\cite{barth2018distributed}. We used open source implementations from OpenAI and Dopamine\footnote{https://github.com/openai/baselines, https://github.com/google/dopamine.}. For DDPG training, we used default parameters, but ensured that the model architecture, batch size, exploration noise, number of updates and other parameters matched that of the WCPG training parameters. PPO training was performed with minibatch size of 32, $\gamma=0.99$, $\lambda=0.95$, $clip=0.2$, and learning rate of $0.0003$. Training was performed until convergence.

For all of the methods, \textit{\textbf{training}} performance is quite good with almost all of them achieving $0\%$ crash rates. However, if we use \emph{out-of-distribution} environment parameters, PPO shows severe overfitting. DDPG is better but also starts to experience collisions. Note that C51/Rainbow and D4PG are \emph{distributional} RL techniques (highlighted in blue), but they do not optimize for any risk-sensitive criteria. For all baseline methods, the performance significantly degrades in the presences of changing environmental parameters. In contrast, WCPG policies are more robust as we reduce $\alpha$, achieving $0\%$ crash rate for the left turn environment and between $0\%$ and $1\%$ for the merge environment, while keeping the success rate high. See Appendix for full results.

\begin{figure*}[t]
\hspace{-0.05in}
    \begin{subfigure}[c]{0.18\textwidth}
        \includegraphics[width=\textwidth]{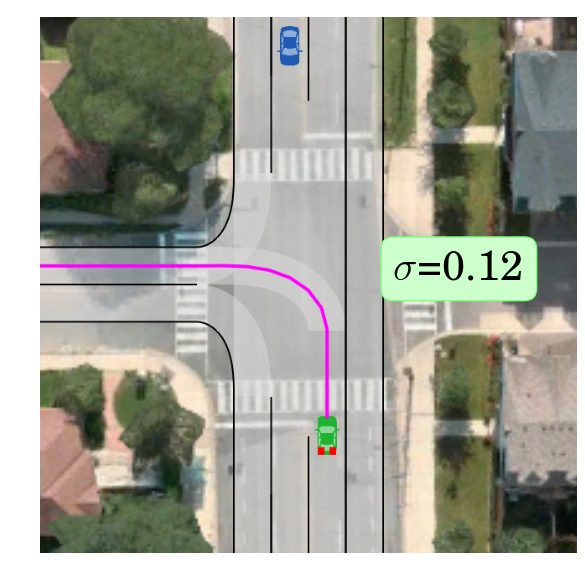}\vspace{-0.1in}
        \caption{t=0.0 s}
    \end{subfigure}     \hspace{-0.15in}
    \begin{subfigure}[c]{0.18\textwidth}
        \includegraphics[width=\textwidth]{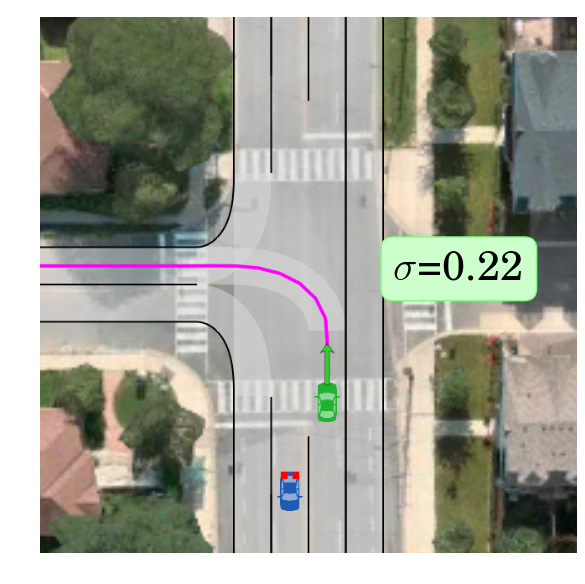}\vspace{-0.1in}
        \caption{t=4.8 s}
    \end{subfigure}    \hspace{-0.15in}
    \begin{subfigure}[c]{0.18\textwidth}
        \includegraphics[width=\textwidth]{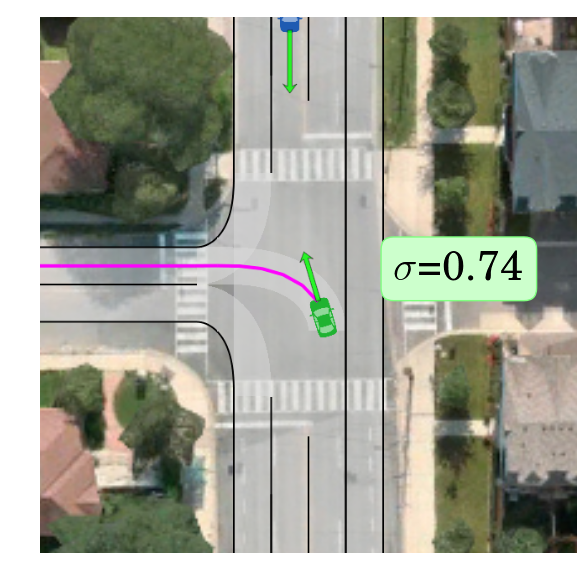}\vspace{-0.1in}
        \caption{t=6.0 s}
    \end{subfigure} \hspace{-0.15in}
    \begin{subfigure}[c]{0.18\textwidth}
        \includegraphics[width=\textwidth]{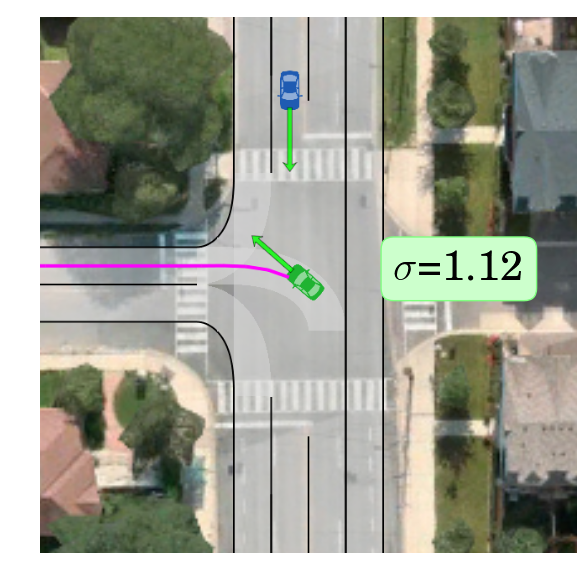}\vspace{-0.1in}
        \caption{t=6.4 s}
    \end{subfigure} \hspace{-0.15in}
        \begin{subfigure}[c\vspace{-0.05in}]{0.18\textwidth}
        \includegraphics[width=\textwidth]{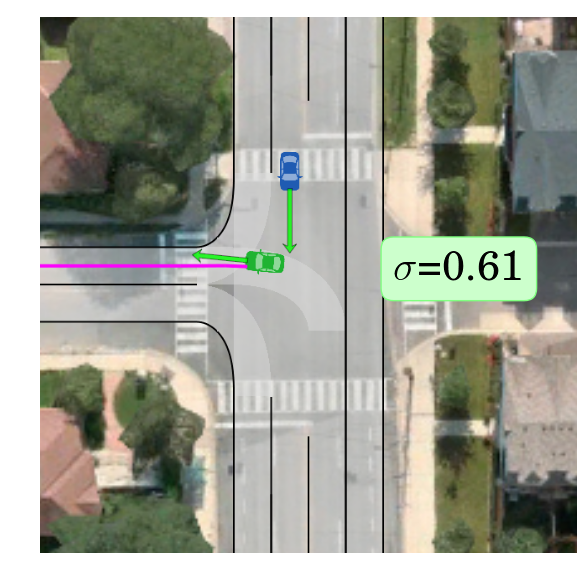}\vspace{-0.1in}
        \caption{t=6.8 s}
    \end{subfigure} \hspace{-0.15in}
    \begin{subfigure}[c]{0.18\textwidth}
        \includegraphics[width=\textwidth]{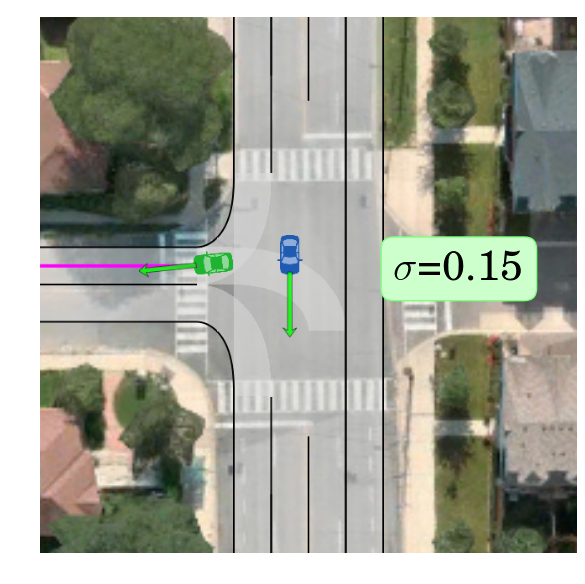}\vspace{-0.1in}
        \caption{t=7.2 s}
    \end{subfigure} \hspace{-0.15in}
    \vspace{-0.1in}
    \caption{\small {Critic's estimation of the standard deviation of the future return. Uncertainty increases dramatically as ego goes ahead of an oncoming vehicle and reduces as ego completes the maneuver. Variance is due to the fact that the behavior of the oncoming vehicles can be random (yield/ignore/accelerate). See text for details.}}
    \vspace{-0.1in}
\label{fig:visualize_uncertainty}
\end{figure*}

\vspace{-0.1in}
\subsection{Uncertainty Modeling}\vspace{-0.1in}
It is also interesting to examine how the critic predicts uncertainty by looking at the critic's own estimate of the standard deviation of future return. Fig.~\ref{fig:visualize_uncertainty} shows six intermediate steps from a test episode along with critic's estimations. As ego approaches the intersection and begins to make the left turn, the variance gradually increases and crescendos when ego is in the path of oncoming traffic. As ego completes the turn, uncertainty quickly reduces.

\vspace{-0.1in}
\subsection{Carla Simulation}
\begin{figure}
\begin{minipage}[c]{0.6\textwidth}
  \includegraphics[width=1\textwidth,height=0.2\textheight]{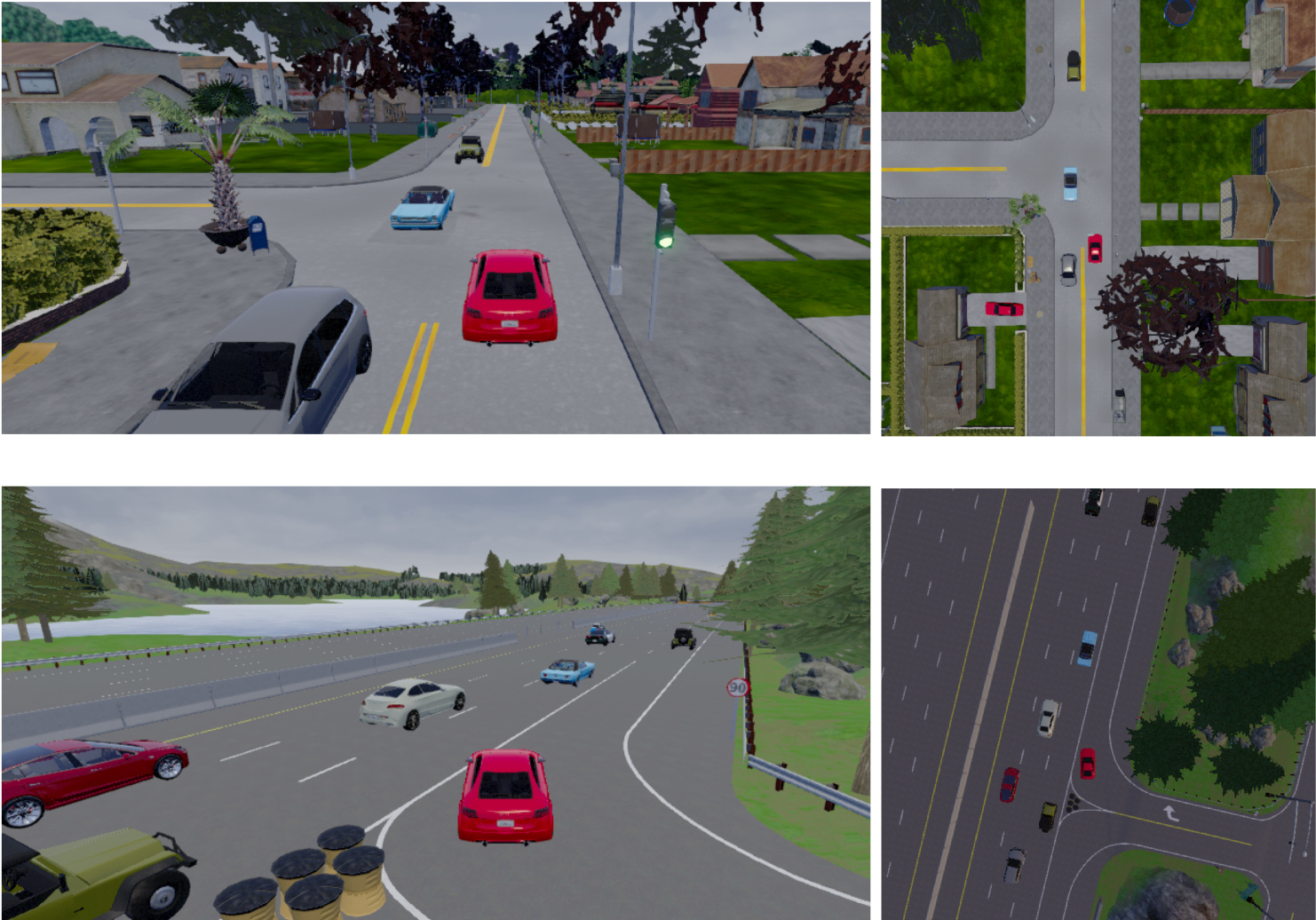}
  \captionof{figure}{\small {CARLA scenarios. Left: 3D view. Right: top-down view.}}
  \label{fig:carla}
\end{minipage}
\hspace{0.1in}
\vspace{-0.2in}
\begin{minipage}[c]{0.37\textwidth}

\vspace{-0.1in}
  \setlength{\tabcolsep}{2pt} 
  \captionof{table}{\small {Collision and (success rates) for different $\alpha$ in CARLA scenarios.} }
\makebox[1\textwidth][c]{
    \begin{tabular}{lccc}
\toprule
\multicolumn{4}{c}{Unprotected Left Turn: (Town05) } \\
 & \multicolumn{1}{c}{$\alpha\!=\!0.2$} & \multicolumn{1}{c}{$\alpha\!=\!0.5$} & \multicolumn{1}{c}{$\alpha\!=\!1.0$} \\
\midrule
& \s{0}{100} & \s{24}{76}  & \s{42}{58}  \\
\bottomrule
\toprule
\multicolumn{4}{c}{Merge: (Town04) } \\
 & \multicolumn{1}{c}{$\alpha\!=\!0.2$} & \multicolumn{1}{c}{$\alpha\!=\!0.5$} & \multicolumn{1}{c}{$\alpha\!=\!1.0$} \\
 \midrule
& \s{2}{83} & \s{4}{89}  & \s{24}{76}  \\
\bottomrule
\end{tabular}%
}
\label{tab:carla}
\end{minipage}
\end{figure}

\vspace{-0.1in}
We tested how well our policy can transfer to similar scenarios in the CARLA simulator~\cite{Dosovitskiy17}. It currently contains seven different towns, dozens of different vehicles, and simple traffic law abiding ``auto-pilot" CPU agents. For our left turn and merge environments, we found locations within CARLA that had similar geometry (See Fig.~\ref{fig:carla}). We extract the position, heading, and velocity of vehicles from the CARLA simulator and fed into our previously trained WCPG policy networks\footnote{Due to a much slower simulation speed, we leave training directly on CARLA as future work.}. We report collision and success rates for 100 random trials in Tab.~\ref{tab:carla}. We can see that even in a different simulation, lower $\alpha$ values improved robustness dramatically.

\vspace{-0.1in}
\section{Related Work}\vspace{-0.15in}
Risk-sensitive, safe RL and the related robust MDPs have been extensively studied in literature~\cite{bagnell2001solving,NilimE05,MorimotoD05,PintoDSG17}. A recent comprehensive survey on the topic is provided by~\cite{GarciaF15}, where safe RL is categorized into two main types. The first type is based on the modification of the exploration process to avoid unsafe exploratory actions. Strategies consist of incorporating external knowledge~\cite{GeibelW05} or using risk-directed explorations~\cite{gehring2013smart,Law05}. The second type modifies the optimality criterion~\cite{howard1972risk,Heger94,MihatschN02,Borkar02,MorimuraSKHT10} used during training. Our work falls under this latter category, where we try to optimize our policy to strike a balance between pay-off and avoiding catastrophic events.

Different types of ``safe" criteria have been proposed: exponential utility functions~\cite{howard1972risk}, linear combination of return and variance~\cite{SatoEtAl01}, percentile performance~\cite{filar1995percentile}, Sharpe ratio and other variance-related criteria~\cite{TamarCM12}. The worst-case minimax criterion can also be directly maximized~\cite{Heger94,littman1996generalized}. In~\cite{Heger94}, $\hat{Q}$-Learning was introduced, where the $\hat{Q}$ function is the lower bound of the $Q$ function. However, optimizing the minimax criterion can result in overly pessimistic policies~\cite{Gaskett03}. Optimizing a constrained criterion (return) subjected to a bounded variance was proposed in~\cite{PrashanthG13}.

Conditional Value-at-Risk (CVaR) is another criterion that is gaining popularity in various fields such as engineering and finance~\cite{rockafellar2000optimization, chow2015risk}. It has various desirable properties, including coherence and the easy of computation for certain underlying distributions. Combing CVaR with reinforcement learning, policy gradients for a bounded CVaR was proposed  by~\cite{chow2014algorithms}, while a sampling based algorithm for CVaR was discussed in~\cite{TamarGM15}. However, these approaches directly estimate the gradients to CVaR and are often computationally expensive or require extensive trajectory roll-outs. In contrast, WCPG optimizes for CVaR \emph{indirectly} by first using \emph{distributional} RL techniques~\cite{jaquette1973markov,Sobel82,white1988mean} to estimate the distribution of return and then compute CVaR from this distribution. Recently, there have been a resurgence of interest in distributional RL~\cite{BellemareDM17,dabney2017distributional,barth2018distributed}. However, the estimated distributions have not been used to minimize any risk-sensitive criteria. While implicit quantile networks~\cite{DabneyOSM18} do optimize for risk-sensitive measures, a disadvantage to their approach is the approximation of risk and its gradients via $K$ discrete samples, which greatly increases computation complexity.

Our proposed WCPG builds on DDPG, a popular deep RL method for continuous control. We retain DDPG's advantage of a powerful and sample efficient off-policy method that works well for large continuous state and action space. Instead of directly optimizing for CVaR, which can be difficult, we compute CVaR in closed-form from our distributional critic's estimation of future return, without resorting to sampling. In addition, both our actor and critic take risk-tolerance $\alpha$ as input during training, which allows the learned policy to operate with varying levels of risk after training.

\vspace{-0.1in}
\section{Discussions}\vspace{-0.15in}
We have proposed a novel actor-critic framework to learn risk-sensitive policies by maximizing CVaR. Our policies can be adjusted dynamically after deployment to select risk-sensitive actions. In simulated driving environments, we perform significantly better with a smaller $\alpha$, when compared to other top RL algorithms. Modeling uncertainty with heavy-tailed or mixture distributions could be explored in the future.
It would also be interesting to see if the automatic selection of $\alpha$ would be possible for completing a maneuver while minimizing risk exposure.

\vspace{0.1in}
\noindent
\textbf{Acknowledgements}
\ \ We thank Barry Theobald, Hanlin Goh, Nitish Srivastava, Johannes Heinrich, and the anonymous reviewers for making this a better manuscript.
\vspace{-0.1in}

\small
\bibliography{wcpg}  

%
\clearpage
\appendix
\section*{APPENDIX}
\section{Proof of Proposition 1}
For convenience, we first restate the definitions here:
\[
Q^\pi(s,a) = \mathbb{E} \big[ \sum^\infty_{t=0} \gamma^{t} r(s_t,a_t) | s_0=s, a_0=a, \pi \big ]
\]
\[
\Upsilon^\pi(s,a) = \mathbb{E} \big [ \big (\sum^\infty_{t=0} \gamma^{t} r(s_t,a_t) \big)^2 | s_0=s, a_0=a, \pi \big ] - Q^\pi(s,a)^2
\]
For conciseness, we define $Q_{sa}^\pi \doteq Q^\pi(s,a)$, $\Upsilon_{sa}^\pi \doteq \Upsilon^\pi(s,a)$, $R_{sa}^t \doteq r(s_t,a_t)$, $ \mathbb{E}^{\pi}_{sa}[ \sum^\infty_{t=0} \gamma^t R_{sa}^t ]       \doteq     \mathbb{E} \big [ \big (\sum^\infty_{t=0} \gamma^{t} r(s_t,a_t) \big)^2 | s_0=s, a_0=a, \pi \big ]  $.
We also use $s', a'$ to denote the next resulting state and action obtained from taking action $a$ in state $s$, and $a' \leftarrow \pi(s')$.

\begin{proof}
\begin{align}
\Upsilon_{sa}^\pi 	&= \mathbb{E}^{\pi}_{sa} \Big [ \big( \sum^\infty_{t=0} \gamma^t R_{sa}^t \big )^2 \Big ]  - ( Q_{sa}^\pi )^2 \\
					&= \mathbb{E}^{\pi}_{sa} \Big [ \big( R_{sa}^0 + \sum^\infty_{t=1} \gamma^t R_{s'a'}^t \big )^2 \Big ]  - ( Q_{sa}^\pi )^2
\end{align}
We now take out the first term out of the summation:
\small
\begin{align}
\Upsilon_{sa}^\pi   &= (R_{sa}^0)^2 + 2 \gamma^t R_{sa}^0 \sum_{s'} p(s'|s) \mathbb{E}^{\pi}_{s'a'} \Big [ \sum^\infty_{t=1} \gamma^t R_{s'a'}^t \Big ] + \sum_{s'} p(s'|s) \mathbb{E}^{\pi}_{s'a'} \Big [ \big ( \sum^\infty_{t=1} \gamma^t R_{s'a'}^t \big )^2 \Big ] - ( Q_{sa}^\pi )^2
\end{align}
\normalsize
Now, we can reuse the definition of $Q_{sa}^\pi$ and $\Upsilon_{sa}^\pi$ to arrive at the recursion:
\small
\begin{align}
\Upsilon_{sa}^\pi  &= (R_{sa}^0)^2 + 2 \gamma^t R_{sa}^0 \sum_{s'} p(s'|s) Q_{s'a'}^\pi + \gamma^2 \sum_{s'} p(s'|s)\Upsilon_{s'a'}^\pi + \gamma^2 \sum_{s'} p(s'|s) Q_{s'a'}^\pi - ( Q_{sa}^\pi )^2
\end{align}
\normalsize
\end{proof}
\section{Proof of Theorem 2}

\subsection*{Condition A.1}
The policy function $\pi$ is deterministic: $\pi(a|s) \triangleq \delta(a')$, where $a' \leftarrow \pi(s)$ and $\delta(\cdot)$ is the Dirac delta function. This is a mild condition as our proposed WCPG is based on the deterministic policy gradient (DDPG), which also assumes deterministic policy functions.
\subsection*{Condition A.2}
Given a particular state $s$ and action $a$, the environment transition is deterministic, that is:
$p(s'|s,a) \triangleq \delta(s')$, where $\delta(\cdot)$ is the Dirac delta function. We note here that environment determinism for continuous control is often made by widely popular RL environments. For example, both Mujoco simulator and the DeepMind Control Suite are deterministic. The original Atari Learning Environment (ALE) also has deterministic dynamics.

For clarity and without loss of generality, we define $\Gamma$ without conditioning on risk parameter $\alpha$:
\begin{align}
\Gamma^\pi(s,a,\alpha) &= Q^\pi(s,a) -  \frac{\phi(\alpha)}{\Phi(\alpha))} \sqrt{ \Upsilon^\pi(s,a) } \\
\Gamma^\pi_{sa}					   &= Q^\pi_{sa} - C \sqrt{\Upsilon^\pi_{sa} },
\end{align}
where $C$ is an $\alpha$ dependent constant.

\begin{customprop}{3}
Let us define $\Gamma^\pi_{s}$ to be the expected conditional Value-at-Risk for policy $\pi$ starting from state $s$, and given Condition A.1 is satisfied,
\begin{equation}
\Gamma^\pi_{s} = \sum_a \pi(a|s) \Gamma^\pi_{sa}.
\end{equation}
Proof. It is trivial to see that as $\pi$ approach the Dirac delta function, probability mass concentrate on the chosen deterministic action $a$.
\end{customprop}

\begin{customprop}{4}
Provided that Condition A.2 is satisfied,
\begin{equation}
\Gamma^\pi(s,a) = r(s,a) + \gamma \sum_{s',a'} p(s'|s,a) \Gamma^\pi(s', a')
\end{equation}
\begin{proof}
\begin{align}
\Gamma^\pi_{sa}		&= Q^\pi_{sa} - C \sqrt{\Upsilon^\pi_{sa} } \\
					&= \big ( r_{sa} +\gamma \sum_{s',a'} p(s'|s,a) Q^\pi_{s'a'} \big ) - C \sqrt{\Upsilon^\pi_{sa} }
\end{align}
From Eq. 4 in proof of Proposition 1, we substitute $\Upsilon^\pi_{sa}$:
\small
\begin{align}
\Gamma^\pi_{sa}  	&= \big ( r_{sa} +\gamma \sum_{s',a'} p(s'|s,a) Q^\pi_{s'a'} \big ) \\ 
					&-C \sqrt{ 
(R_{sa}^0)^2 + 2 \gamma R_{sa}^0 \sum_{s'} p(s'|s) Q_{s'a'}^\pi + \gamma^2 \sum_{s'} p(s'|s)\Upsilon_{s'a'}^\pi + \gamma^2 \sum_{s'} p(s'|s) Q_{s'a'}^\pi - ( Q_{sa}^\pi )^2
}.
\end{align}
Given Condition 2, we simplify expectations involving summations over $s'$, and rearranging:
\begin{align}
\Gamma^\pi_{sa}  	&= \big ( r_{sa} +\gamma \sum_{s',a'} p(s'|s,a) Q^\pi_{s'a'} \big )
					-C \sqrt{ 
(R_{sa}^0)^2 + 2 \gamma^t R_{sa}^0 Q_{s'a'}^\pi + \gamma^2 \Upsilon_{s'a'}^\pi + \gamma^2 Q_{s'a'}^\pi - ( Q_{sa}^\pi )^2
} \\
&= \big ( r_{sa} +\gamma \sum_{s',a'} p(s'|s,a) Q^\pi_{s'a'} \big ) -C \sqrt{ 
\gamma^2 \Upsilon_{s'a'}^\pi + ( R_{sa}^0 + 2 \gamma Q_{s'a'}^\pi )^2 - ( Q_{sa}^\pi )^2
} \\
&= \big ( r_{sa} +\gamma \sum_{s',a'} p(s'|s,a) Q^\pi_{s'a'} \big ) -C \sqrt{ 
\gamma^2 \Upsilon_{s'a'}^\pi + ( Q_{sa}^\pi )^2 - ( Q_{sa}^\pi )^2
} \\
&= r_{sa} +\gamma \sum_{s',a'} p(s'|s,a) Q^\pi_{s'a'} - \gamma C \sqrt{ 
 \Upsilon_{s'a'}^\pi }  \\
 &= r_{sa} +\gamma \sum_{s',a'} p(s'|s,a) \big ( Q^\pi_{s'a'} - C \sqrt{ \Upsilon_{s'a'}^\pi }  \big )  \\
 &= r_{sa} +\gamma \sum_{s',a'} p(s'|s,a) \Gamma^\pi_{s'a'}
\end{align}
\end{proof}
\end{customprop}
\normalsize
\subsubsection*{Proof of Theorem 2.}
\begin{proof}
We mainly follow the start state formulation of the Policy Gradient Theorem of (Sutton et al. 2000), but also making use of Propositions 3 and 4.
\begin{align}
\frac{\partial \Gamma^\pi_{s} }{\partial \theta } &\doteq \frac{\partial}{\partial \theta } \sum_a \pi(a|s) \Gamma^\pi_{sa}, ~~~~~  \forall s \in \mathcal{S} \mbox{\hspace{2.in}(Proposition 3)}\\
&= \sum_a \Bigg [ \frac{\partial  \pi(a|s) }{\partial \theta } \Gamma^\pi_{sa} +  \pi(a|s) \frac{\partial \Gamma^\pi_{sa}}{\partial \theta } \Bigg ] \\
&= \sum_a \Bigg [ \frac{\partial  \pi(a|s) }{\partial \theta } \Gamma^\pi_{sa} +  \pi(a|s) \frac{\partial }{\partial \theta } \Big[ r_{sa} +\gamma \sum_{s',a'} p(s'|s,a) \Gamma^\pi_{s'a'} \Big] \Bigg ] \mbox{\hspace{0.2in}(Proposition 4)} \\
&= \sum_a \Bigg [ \frac{\partial  \pi(a|s) }{\partial \theta } \Gamma^\pi_{sa} +  \pi(a|s) \gamma \sum_{s',a'} p(s'|s,a) \frac{\partial \Gamma^\pi_{s'a'}}{\partial \theta } \Bigg ] \\
&= \sum_x \sum_t^\infty \gamma^t Pr(s \rightarrow x, t, \pi) \sum_a \frac{\partial \pi(a|x)}{ \partial \theta} \Gamma^\pi_{x,a}, \nonumber
\end{align}
where we have used $Pr(s \rightarrow x, k, \pi)$ to denote the probability of going to state $x$ from state $s$ under policy $\pi$ in $k$ steps. This can be seen after unrolling (Eq. 22) for a few steps. It can then be seen that:
\begin{align}
\frac{\partial J}{ \partial \theta} &= \frac{\partial \Gamma^\pi_{s_0} }{\partial \theta }, \mbox{ for some initial state $s_0$} \\
&= \sum_s \sum_{t=0}^\infty \gamma^t Pr(s_0 \rightarrow s, t, \pi) \sum_a \frac{\partial \pi(a|s)}{ \partial \theta} \Gamma^\pi_{s,a} \\
&= \sum_s \rho(s) \sum_a \frac{\partial \pi(a|s)}{ \partial \theta} \Gamma^\pi_{s,a}
\end{align}
Using the log-derivative trick we can further write the gradient as:
\begin{align}
\frac{\partial J}{ \partial \theta} &= \sum_s \rho(s) \sum_a \pi_\theta(a|s) \nabla_\theta \log \pi_\theta(a|s) \Gamma^\pi(s,a)  \\
\end{align}
And for conditional objective conditioned on $\alpha$, we have:
\begin{align}
\frac{\partial J_\alpha}{ \partial \theta} &= \sum_s \rho(s) \sum_a \pi_\theta(a|s, \alpha) \nabla_\theta \log \pi_\theta(a|s, \alpha) \Gamma^\pi(s,a, \alpha)
\end{align}
\end{proof}

\section{Algorithm}
\begin{algorithm}[H]
\small
\DontPrintSemicolon
\SetAlgoLined
\SetKwInput{Input}{Input~}
\SetKwInput{Initialize}{Initialize~}
\Input{~Environment $\mathcal{E}$, actor-critic net hyperparameters, $M$ episodes, T steps\slash episode, replay buffer $\mathcal{B}$.}
\Initialize{Randomly initialize actor $\pi$ and critic, $\mathcal{B} \leftarrow \emptyset$.}
\For{episodes $e = 0 \mbox{ to } M$}{
	$s_0 \leftarrow \mathcal{E}$.reset(), $\alpha \sim U(0,1)$ \;
    \For{steps $t = 0 \mbox{ to } T$}{
      action $a_t \leftarrow \pi_\alpha(s_{t})$; add exploration noise. \;
      $\lbrace s_{t+1}, r_t, done \rbrace \leftarrow \mathcal{E}$.step($a_t$) \;
      $\mathcal{B} \leftarrow \lbrace s_t,a_t,r_t,s_{t+1},\alpha  \rbrace  $ \;
      Randomly sample a minibatch from $\mathcal{B}$. \;
      Update critic: use Eqs.~\ref{eq:dis_bell} and ~\ref{eq:wasserstein}. \;
      Update actor with the CVaR objective Eq.~\ref{eq:dwcpg}. \;
      \lIf{$done$}{goto next episode.}
    }    
}
\caption{WCPG training algorithm}
\label{alg:wc_pg}
\end{algorithm}

\section{Distributional Critic}
Our distributional critic uses the Gaussian approximation as it allows us to have closed form CVaR computation, which is critical as we must compute CVaR for every update step, for every data tuple. In addition, ease of computation is important when we want to learn a policy which is conditioned on the risk $\alpha$, where we have to compute CVaR for multiple $\alpha$s.

While a Gaussian is unimodal and not heavy-tailed, our environmental reward is lower-bounded by the negative reward caused by a collision. In this case, Gaussian distribution's light-tail is not terribly restrictively, since we do not have unbounded loss. In addition, it is important to note that the unimodal Gaussian approximation is for a particular state-action pair. The value distribution for a particular state $s$ (e.g. distributional $V(s)$), which marginalizes over the actions, could still be multi-modal with even potentially different variances for different modes.

\section{Experiments}
\subsection{Driving Simulation}
We focused our experiments on self-driving environments as safety and risk-sensitive decision making is paramount in this application. They are also multi-agent environments, where the latent behaviors of other agents are not always observable. Specifically, as noted in the paper, the other on-coming vehicles have a random probability of 3 behaviors: yielding, ignoring, or aggressively accelerate. This stochasticity creates inherent uncertainty in the environment and leads to the need for a distributional RL.

\subsubsection{Driving simulation experiment details}
See Table~\ref{tab:appendix_driving_env} for the details of the training environment parameters.
\begin{table}[h!]
  \centering
  \setlength{\tabcolsep}{3pt}
  \caption{Driving Environment Details}
  \label{tab:appendix_driving_env}
\scriptsize
  \begin{tabular}{cccccccc}
    \toprule
    Environment & Scale (m) & Ego vel. (m/s) & Agents vel. & Spawn rate & Yield/ignore/aggressive & $R_{success}$ & $R_{collision}$ \\
    \midrule
    Unprotected Turn & 240 & [0, 20] & [10, 20]& 1\% & 0\%, 80\%, 20\% & $50 \times e^{-steps/50}+10$ & -50.0 \\
    \midrule
    Merge 		    & 180 & [5, 20] & [5, 15] & 1\% & 20\%, 40\%, 40\% & $75 \times e^{-steps/50}+5$ & -50.0 \\
  \bottomrule
\end{tabular}
\end{table}

\subsubsection{Network Parameters}
\begin{table}[h]
\centering
  \setlength{\tabcolsep}{4pt} 
  \caption{WCPG actor and critic networks for driving environments.}
  \label{tab:network_params}
  \begin{tabular}{cccccccccc}
    \toprule
     \multicolumn{5}{c}{Actor Network}  & \multicolumn{5}{c}{Critic Network} \\
     \cmidrule(lr){1-5}  \cmidrule(lr){6-10}
    layer & dim. & prev. & acti. & $\sigma$ &  layer & dim. & prev. & acti. & $\sigma$ \\
     \cmidrule(lr){1-5}  \cmidrule(lr){6-10}
     $I$  &  16 & -  & linear &  - 				
        & $I$                &  16 & -  & linear &  -	\\
     $I_\alpha$ &  1 & -& linear & -  							
      	& $I_{act}$ &  1 & -& linear & -       			    \\
     $h_1$ & 32 & $I$ & ReLU & $\sqrt{2}$ 					
     	& $I_\alpha$ &  1 & -& linear & -               \\
     $h_2$ & 16 & $I_\alpha$ & ReLU & $\sqrt{2}$ 			
        & $h_1$ & 64 & $I$ & ReLU & 0.01               \\
     $h_3$ & 32 & $h_1, h_2$ & ReLU & $\sqrt{2}$ 				
        & $h_2$ & 64 & $I_\alpha$ & ReLU & 0.01     \\
     $O_{act}$ & 1 & $h_3$ & tanh & $\sqrt{2}$   					
        & $h_3$ & 64 & $h_1, h_2, I_{act}$ & ReLU & 0.01     \\
     &  &  &  &    
        & $h_4$ & 64 & $h_3$ & ReLU & 0.01 \\  
     &  &  &  &  
        & $O_{critic}$ & 2 & $h_4$ & linear & 0.01 \\
     &  &  &  &  
        &  &  &    & softplus & 0.01 \\
    \midrule 
    \bottomrule
  \end{tabular}
\end{table}

We show the exact network parameters for the WCPG actor and critic network used for the driving environment simulations in Table~\ref{tab:network_params}.

\subsection{Additional Proof-of-concept Experiment}
\subsubsection{Fast-Slow Lane Selection}
We demonstrate the concept of WCPG in a simple discrete setting where we must learn a policy to decide to drive in the fast lane or the slow lane on a highway. The problem is a highly simplified proof-of-concept MDP where we can leverage brute-force sampling to perform policy improvement and find the optimal policy. This allows us to test the core concepts behind WCPG independently from the various approximations needed in learning deep RL policies for more complex environments.
\begin{figure}[h!]
\begin{center}
\includegraphics[width=0.7\linewidth]{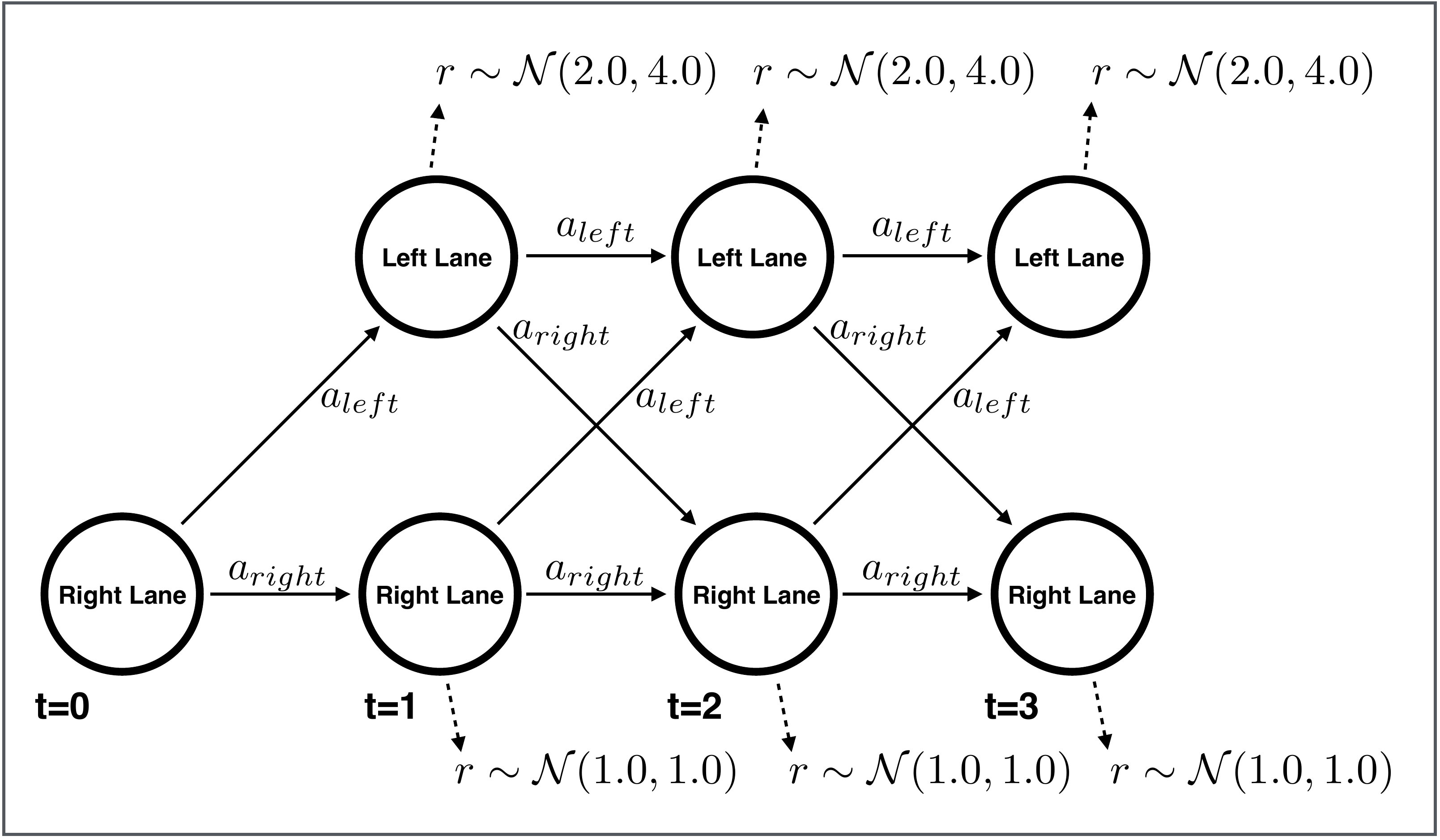}
\end{center}
\vspace{-0.1in}
\caption{{MDP for the Fast-Slow Lane Selection problem. The goal is to learn a policy for choosing either the left(fast) or the right(slow) lane, see text for details. This proof-of-concept problem demonstrate the core concepts of WCPG.}}
\label{fig:left_right_lanes}
\end{figure}
Imagine a vehicle is traveling on a two lane freeway and needs to learn a policy for performing lane \emph{selection}. The left lane is the fast lane while the right lane is the slower lane. At every timestep, the rewards are stochastic and the fast lane reward has both higher mean and higher variance than the slow lane. The stochastic policy we wish to learn is the probability of lane change at each timestep.

We formulate this as a discrete finite-horizon MDP with two states, $s_{left}$, $s_{right}$ for the vehicle being in the left and right lanes, respectively. The MDP is shown in Fig.~\ref{fig:left_right_lanes}, where the total number of time steps is 4. The initial state at $t=0$ is the right lane and at each timestep the policy outputs the probability of lane change: $Pr(left \ lane) = 1.0 - Pr( right \ lane) $. When the vehicle is in the left lane, it receives a reward randomly sampled from $\mathcal{N}(2.0, 4.0) $. When the vehicle in the right lane, its reward is sampled from $\mathcal{N}(1.0, 1.0) $. This roughly models the assumption of higher risk and rewards for the fast lane, while the slow lane might be safer but would take longer to get to the destination.

Using this environment, we are interested in learning an $\alpha$ conditioned optimal policy and to verify our hypothesis that different $\alpha$s will lead to different policies. The MDP is simple enough such that sampling based policy improvement schemes will arrive at the optimal solution.

\begin{figure}[t]
\begin{center}
\includegraphics[width=0.8\textwidth]{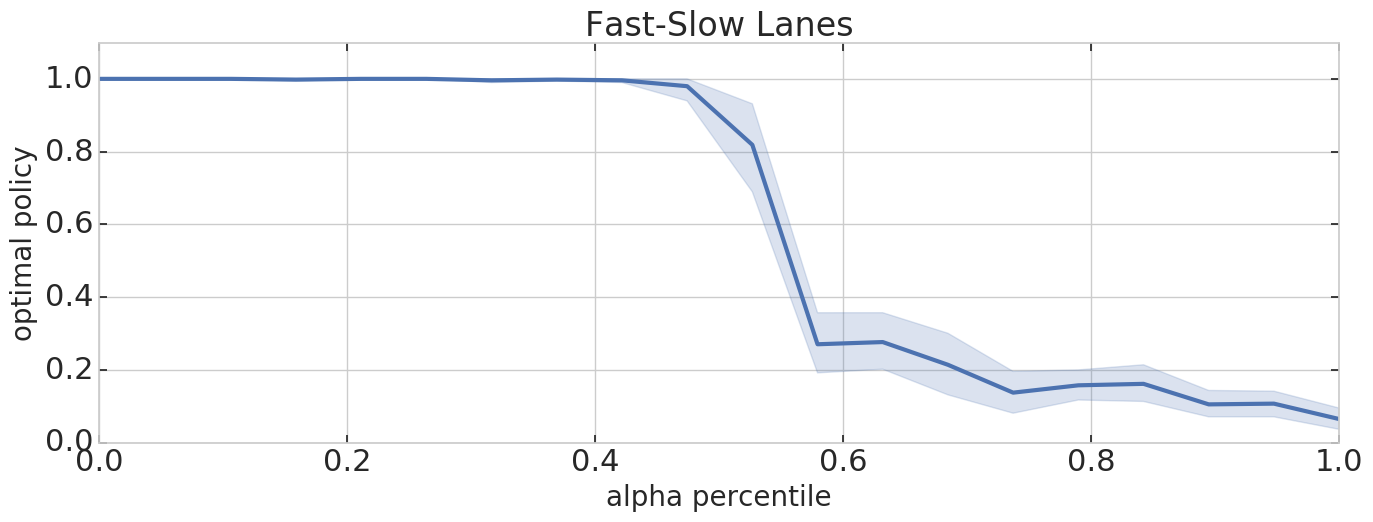}
\end{center}
\vspace{-0.15in}
\caption{{Optimal policy as a function of $\alpha$ percentile.}}
\label{fig:fast_slow_lanes}
\vspace{-0.1in}
\end{figure}

The Fast-Slow Lanes MDP can be solved via brute force sampling based policy improvement. For a given $\alpha$, we grid the entire policy space. Specifically, we enumerate $\alpha$ from $0.0$ to $1.0$ with 32 intervals. Iterating through the policies, we numerically evaluate each policy using 1000 random trials, where the CVaR objective is computed from the statistics of the trials. The optimal policy is then found by selecting the policy that achieves the highest CVaR for any given $\alpha$. We plot the optimal policies in Fig.~\ref{fig:fast_slow_lanes}.

The optimal policy is a scalar specifying the probability for left/right lane selection. When $\alpha$ is small, the optimal policy is to stay in the right lane as the rewards have a much smaller variance. Conversely, when $\alpha$ approaches $1.0$, the optimal policy is to choose the left lane as we obtain higher rewards in expectation. This results verifies our hypothesis and is expected given the specified distribution of rewards. While this is a simple proof-of-concept problem, it demonstrates that when a globally optimal solution can be found, the proposed CVaR objective does indeed lead to conditional policies that varies depending on user specified risk appetite.

\subsection{Extrapolation Experiments Full Results}\label{sec:extrap_full_results}
We report the full extrapolation results for the unprotected left and merge environments in Tables~\ref{tab:full_unp_left_collision},~\ref{tab:full_unp_left_success},~\ref{tab:full_merge_collision},~\ref{tab:full_merge_success}.
The numbers are all in percentages and we also report the standard errors of the mean, over 100 trials with random environment seeds.
\begin{figure}[h]
\begin{minipage}[t]{1.0\textwidth}
\small
  \setlength{\tabcolsep}{3pt} 
  \captionof{table}{{ Unprotected Left Turn (extrapolation) - Collision rate $\%$}} \vspace{-0.1in}
\makebox[1\textwidth][c]{
    \begin{tabular}{rr >{\columncolor{lightgray2}}r >{\columncolor{lightgray2}}r >{\columncolor{lightgray2}}r >{\columncolor{lightgray2}}r >{\columncolor{lightgray2}}r r r>{\columncolor{blue}}r >{\columncolor{blue}}r }
\toprule
 \multicolumn{2}{c}{Params}  & \multicolumn{5}{c}{WCPG (varying $\alpha$)} & \multicolumn{4}{c}{State-of-the-art baselines} \\
\cmidrule(lr){1-2}  \cmidrule(lr){3-7} \cmidrule(lr){8-11}
vel. & spwn. &   0.02  & 0.1 & 0.3  & 0.6 & 1.0 & DDPG & PPO & C51 & D4PG \\ 
\midrule
(train)&  1\%&	\e{0.0}{0.0}  & \e{0.0}{0.0}  & \e{0.0}{0.0}  & \e{2.0}{1.4} & \e{15.0}{3.6} & \e{0.0}{0.0}  & \e{4.0}{2.0}  & \e{2.0}{1.4} &  \e{2.0}{1.4} \\
\midrule
+5 m/s&   5\%&	\e{0.0}{0.0}  & \e{0.0}{0.0}  & \e{0.0}{0.0}  & \e{0.0}{0.0} & \e{15.0}{3.6} & \e{13.0}{3.4} & \e{24.0}{4.3} & \e{10.0}{3.0} & \e{4.0}{2.0} \\ 
+10 m/s&  5\%&	\e{0.0}{0.0}  & \e{0.0}{0.0}  & \e{0.0}{0.0}  & \e{0.0}{0.0} & \e{15.0}{3.6} & \e{18.0}{3.9} & \e{40.0}{4.9} & \e{11.0}{3.1} & \e{4.0}{2.0} \\
+15 m/s&  5\%& 	\e{0.0}{0.0}  & \e{0.0}{0.0}  & \e{0.0}{0.0}  & \e{3.0}{1.7} & \e{21.0}{4.1} & \e{20.0}{4.0} & \e{49.0}{5.0} & \e{14.0}{3.5} & \e{2.0}{1.4} \\
+0 m/s&   2\%&	\e{0.0}{0.0}  & \e{0.0}{0.0}  & \e{0.0}{0.0}  & \e{0.0}{0.0} & \e{19.0}{3.9} & \e{7.0}{2.6}  & \e{15.0}{3.6} & \e{4.0}{2.0} &  \e{5.0}{2.2} \\
+0 m/s&   8\%& 	\e{0.0}{0.0}  & \e{0.0}{0.0}  & \e{0.0}{0.0}  & \e{3.0}{1.7} & \e{26.0}{4.4} & \e{6.0}{2.4}  & \e{15.0}{3.6} & \e{5.0}{2.2} &  \e{1.0}{1.0} \\
+10 m/s&  8\%& 	\e{0.0}{0.0}  & \e{0.0}{0.0}  & \e{0.0}{0.0}  & \e{2.0}{1.4} & \e{23.0}{4.2} & \e{19.0}{3.9} & \e{40.0}{4.9} & \e{11.0}{3.1} & \e{2.0}{1.4} \\
\bottomrule
\end{tabular}%
\label{tab:full_unp_left_collision}
}
\end{minipage}

\begin{minipage}[t]{1.0\textwidth}
\vspace{0.1in}
\footnotesize
  \setlength{\tabcolsep}{3pt} 
  \captionof{table}{{ Unprotected Left Turn (extrapolation) - Success rate $\%$}} \vspace{-0.1in}
\makebox[1\textwidth][c]{
    \begin{tabular}{rr >{\columncolor{lightgray2}}r >{\columncolor{lightgray2}}r >{\columncolor{lightgray2}}r >{\columncolor{lightgray2}}r >{\columncolor{lightgray2}}r r r>{\columncolor{blue}}r >{\columncolor{blue}}r }
\toprule
 \multicolumn{2}{c}{Params}  & \multicolumn{5}{c}{WCPG (varying $\alpha$)} & \multicolumn{4}{c}{State-of-the-art baselines} \\
\cmidrule(lr){1-2}  \cmidrule(lr){3-7} \cmidrule(lr){8-11}
vel. & spwn. &   0.02  & 0.1 & 0.3  & 0.6 & 1.0 & DDPG & PPO & C51 & D4PG \\
\midrule
(train)&  1\%&	\e{100.0}{0.0}  & \e{100.0}{0.0}  & \e{100.0}{0.0}  & \e{98.0}{1.4}  & \e{85.0}{3.6} & \e{100.0}{0.0} & \e{96.0}{2.0} & \e{98.0}{1.4}  & \e{98.0}{1.4} \\
\midrule
+5 m/s&   5\%&	\e{100.0}{0.0}  & \e{100.0}{0.0}  & \e{100.0}{0.0}  & \e{100.0}{0.0} & \e{85.0}{3.6} & \e{87.0}{3.4} & \e{76.0}{4.3} & \e{90.0}{3.0} & \e{96.0}{2.0} \\ 
+10 m/s&  5\%&	\e{100.0}{0.0}  & \e{100.0}{0.0}  & \e{100.0}{0.0}  & \e{100.0}{0.0} & \e{85.0}{3.6} & \e{82.0}{3.9} & \e{60.0}{4.9} & \e{89.0}{3.1} & \e{96.0}{2.0} \\
+15 m/s&  5\%& 	\e{100.0}{0.0}  & \e{100.0}{0.0}  & \e{100.0}{0.0}  & \e{97.0}{1.7}  & \e{79.0}{4.1} & \e{80.0}{4.0} & \e{51.0}{5.0} & \e{86.0}{3.5} & \e{98.0}{1.4} \\
+0 m/s&   2\%&	\e{100.0}{0.0}  & \e{100.0}{0.0}  & \e{100.0}{0.0}  & \e{100.0}{0.0} & \e{81.0}{3.9} & \e{93.0}{2.6} & \e{85.0}{3.6} & \e{96.0}{2.0} & \e{95.0}{2.2} \\
+0 m/s&   8\%& 	\e{100.0}{0.0}  & \e{100.0}{0.0}  & \e{100.0}{0.0}  & \e{97.0}{1.7}  & \e{74.0}{4.4} & \e{94.0}{2.4} & \e{85.0}{3.6} & \e{95.0}{2.2} & \e{99.0}{1.0} \\
+10 m/s&  8\%& 	\e{100.0}{0.0}  & \e{100.0}{0.0}  & \e{100.0}{0.0}  & \e{98.0}{1.4}  & \e{77.0}{4.2} & \e{81.0}{3.9} & \e{60.0}{4.9} & \e{89.0}{3.1} & \e{98.0}{1.4} \\
\bottomrule
\end{tabular}%
}
\label{tab:full_unp_left_success}
\end{minipage}

\begin{minipage}[t]{1.0\textwidth}
\small
\vspace{0.4in}
  \setlength{\tabcolsep}{3pt} 
  \captionof{table}{{ Merge (extrapolation) - Collision rate $\%$}} \vspace{-0.1in}
\makebox[1\textwidth][c]{
    \begin{tabular}{rr >{\columncolor{lightgray2}}r >{\columncolor{lightgray2}}r >{\columncolor{lightgray2}}r >{\columncolor{lightgray2}}r >{\columncolor{lightgray2}}r r r>{\columncolor{blue}}r >{\columncolor{blue}}r }
\toprule
 \multicolumn{2}{c}{Params}  & \multicolumn{5}{c}{WCPG (varying $\alpha$)} & \multicolumn{4}{c}{State-of-the-art baselines} \\
\cmidrule(lr){1-2}  \cmidrule(lr){3-7} \cmidrule(lr){8-11}
vel. & spwn. &   0.02  & 0.1 & 0.3  & 0.6 & 1.0 & DDPG & PPO & C51 & D4PG \\ 
\midrule
(train)&  1\%&	\e{0.0}{0.0}  & \e{0.0}{0.0}  & \e{0.0}{0.0}  & \e{0.0}{0.0} & \e{8.0}{2.7} & \e{0.0}{0.0}  & \e{0.0}{0.0}  & \e{0.0}{0.0} & \e{6.0}{2.4}  \\
\midrule
+5 m/s&   1\%&	\e{0.0}{0.0}  & \e{1.0}{1.0}  & \e{0.0}{0.0}  & \e{2.0}{1.4} & \e{10.0}{3.0} & \e{4.0}{2.0} & \e{31.0}{4.6} & \e{9.0}{2.9}  & \e{9.0}{2.9}  \\ 
+10 m/s&  1\%&	\e{0.0}{0.0}  & \e{0.0}{0.0}  & \e{0.0}{0.0}  & \e{0.0}{0.0} & \e{8.0}{2.7} &  \e{3.0}{1.7} & \e{26.0}{4.4} & \e{5.0}{2.2}  & \e{8.0}{2.7}  \\
+15 m/s&  1\%& 	\e{0.0}{0.0}  & \e{0.0}{0.0}  & \e{0.0}{0.0}  & \e{1.0}{1.0} & \e{9.0}{2.9} &  \e{1.0}{1.0} & \e{24.0}{4.3} & \e{6.0}{2.4}  & \e{9.0}{2.9}  \\
+0 m/s&   2\%&	\e{1.0}{1.0}  & \e{0.0}{0.0}  & \e{1.0}{1.0}  & \e{2.0}{1.4} & \e{18.0}{3.9} & \e{5.0}{2.2} & \e{40.0}{4.9} & \e{8.0}{2.7}  & \e{14.0}{3.5} \\
+0 m/s&   3\%& 	\e{0.0}{0.0}  & \e{0.0}{0.0}  & \e{0.0}{0.0}  & \e{2.0}{1.4} & \e{17.0}{3.8} & \e{5.0}{2.2} & \e{39.0}{4.9} & \e{9.0}{2.9}  & \e{12.0}{3.3} \\
+10 m/s&  3\%& 	\e{0.0}{0.0}  & \e{0.0}{0.0}  & \e{1.0}{1.0}  & \e{2.0}{1.4} & \e{11.0}{3.1} & \e{4.0}{2.0} & \e{33.0}{4.7} & \e{5.0}{2.2}  & \e{10.0}{3.0} \\
\bottomrule
\end{tabular}%
}
\label{tab:full_merge_collision}
\end{minipage}

\begin{minipage}[t]{1.0\textwidth}
\vspace{0.1in}
\footnotesize
  \setlength{\tabcolsep}{3pt} 
  \captionof{table}{{ Merge (extrapolation) - Success rate $\%$}} \vspace{-0.1in}
\makebox[1\textwidth][c]{
    \begin{tabular}{rr >{\columncolor{lightgray2}}r >{\columncolor{lightgray2}}r >{\columncolor{lightgray2}}r >{\columncolor{lightgray2}}r >{\columncolor{lightgray2}}r r r>{\columncolor{blue}}r >{\columncolor{blue}}r }
\toprule
 \multicolumn{2}{c}{Params}  & \multicolumn{5}{c}{WCPG (varying $\alpha$)} & \multicolumn{4}{c}{State-of-the-art baselines} \\
\cmidrule(lr){1-2}  \cmidrule(lr){3-7} \cmidrule(lr){8-11}
vel. & spwn. &   0.02  & 0.1 & 0.3  & 0.6 & 1.0 & DDPG & PPO & C51 & D4PG \\ 
\midrule
(train)&  1\%&	\e{100.0}{0.0}  & \e{100.0}{0.0}  & \e{100.0}{0.0}  & \e{100.0}{0.0} & \e{92.0}{2.7} &\e{100}{0.0} & \e{96.0}{2.0} & \e{100.0}{0.0} & \e{94.0}{2.4} \\
\midrule
+5 m/s&   1\%&	\e{4.0}{2.0}  & \e{5.0}{2.2}  & \e{54.0}{5.0}  & \e{90.0}{3.0} & \e{88.0}{3.3} & \e{96.0}{2.0} & \e{68.0}{4.7} & \e{91.0}{2.9} & \e{91.0}{2.9} \\ 
+10 m/s&  1\%&	\e{2.0}{1.4}  & \e{5.0}{2.2}  & \e{50.0}{5.0}  & \e{93.0}{2.6} & \e{92.0}{2.7} & \e{97.0}{1.7} & \e{73.0}{4.5} & \e{95.0}{2.2} & \e{92.0}{2.7} \\
+15 m/s&  1\%& 	\e{1.0}{1.0}  & \e{5.0}{2.2}  & \e{57.0}{5.0}  & \e{94.0}{2.4} & \e{91.0}{2.9} & \e{99.0}{1.0} & \e{76.0}{4.3} & \e{94.0}{2.4} & \e{91.0}{2.9} \\
+0 m/s&   2\%&	\e{7.0}{2.6}  & \e{11.0}{3.1} & \e{50.0}{5.0}  & \e{78.0}{4.2} & \e{76.0}{4.3} & \e{95.0}{2.2} & \e{59.0}{4.9} & \e{92.0}{2.7} & \e{86.0}{3.5} \\
+0 m/s&   3\%& 	\e{6.0}{2.4}  & \e{10.0}{3.0} & \e{43.0}{5.0}  & \e{70.0}{4.6} & \e{79.0}{4.1} & \e{95.0}{2.2} & \e{59.0}{4.9} & \e{91.0}{2.9} & \e{88.0}{3.3} \\
+10 m/s&  3\%& 	\e{4.0}{2.0}  & \e{8.0}{2.7}  & \e{43.0}{5.0}  & \e{76.0}{4.3} & \e{79.0}{4.1} & \e{96.0}{2.0} & \e{67.0}{4.7} & \e{95.0}{2.2} & \e{90.0}{3.0} \\
\bottomrule
\end{tabular}%
}
\label{tab:full_merge_success}
\end{minipage}

\end{figure}

\end{document}